\documentclass[10pt,letterpaper]{article}
\usepackage[in]{fullpage}

\usepackage{amsmath,amsfonts,bm}

\def\eqref#1{equation~\ref{#1}}

\def\1{\bm{1}}

\DeclareMathAlphabet{\mathsfit}{\encodingdefault}{\sfdefault}{m}{sl}
\SetMathAlphabet{\mathsfit}{bold}{\encodingdefault}{\sfdefault}{bx}{n}

\usepackage{natbib}
\usepackage{amsmath}
\usepackage{amssymb}
\usepackage{booktabs}
\usepackage{graphicx}
\usepackage{multirow}
\usepackage{authblk}
\usepackage{tabularx}
\usepackage{url}
\usepackage{hyperref}
\usepackage[nameinlink,noabbrev]{cleveref}
\usepackage{crossreftools}
\usepackage{etoolbox}
\usepackage{tikz}
\usetikzlibrary{arrows.meta,positioning}
\usepackage{fvextra}
\usepackage{multirow}

\title{LLMAdBench: A Human Preference Benchmark for Advertising in LLM Responses}

\author{
Rui Ai$^{*,1}$,
Yuqing Liu$^{*,2}$,
Sitao Qiu$^{3}$,
Yun Qiao$^{4}$,
Yuhan Wang$^{5}$,
Jessica Xiwen Wang$^{6}$,
Yiqi Yang$^{7}$,
Lihong Huang$^{8}$,
Ruiyao Sun$^{9}$,
Kaifeng Zhang$^{3}$,
Shengze Ding$^{3}$,
Jiaqi He$^{3}$,
Xinman Wang$^{3}$,
Tianhao Gao$^{3}$,
Jimmy Qin$^{10}$,
Jianghao Lin$^{\dagger,3}$,
Chonghuan Wang$^{\dagger,10}$
}
\date{}

\def\compilemode{arxiv}
\begin{document}

\maketitle
\vspace{-1.2cm}
\begin{center}
\small
$^{1}$Massachusetts Institute of Technology;
$^{2}$University of Michigan, Ann Arbor;
$^{3}$Shanghai Jiao Tong University;
$^{4}$The Ohio State University;
$^{5}$University of Hong Kong;
$^{6}$McGill University;
$^{7}$University of Chicago;
$^{8}$Tongji University;
$^{9}$Beijing Normal University;
$^{10}$University of Texas at Dallas
\end{center}


\footnotetext[1]{Equal contribution.}
\footnotetext[2]{Corresponding authors:
\texttt{linjianghao@sjtu.edu.cn},
\texttt{chonghuan.wang@utdallas.edu}}

\begin{abstract}
Inserting advertisements (ads) into consumer-facing LLM output is emerging as a new business model, but there is little shared evidence on how such ad insertion should be evaluated or how it affects user preferences. 
We introduce \textsc{LLMAdBench}, a human-preference benchmark for studying advertising in LLM-generated content. The benchmark isolates a simple but practically important decision: given a user conversation, an LLM response, and a matched advertisement, where should the ad be placed? Our dataset compares pairs of responses that differ only in ad position while holding all other conditions fixed including the user query, base answer, advertisement, and disclosure condition. 
Human annotators evaluate each pair based on six criteria from both advertiser's and user's perspectives. 
The resulting benchmark contains more than 18{,}000 human judgments across two disclosure conditions: explicitly labeling the ad as sponsored and merging it into the response without disclosure. 
We use \textsc{LLMAdBench} to evaluate eight frontier LLMs as preference judges and find that they are not reliable substitutes for human evaluation. Even the most stable models reverse roughly one quarter of their decisions when the presentation order is swapped, agreement across models is low, and their placement preferences differ systematically from those of human annotators. Moreover, \textsc{LLMAdBench} contains substantial learnable signal. In particular, a Qwen3-8B model fine-tuned on the human preferences improves substantially over its base model and outperforms all zero-shot frontier judges on the held-out prediction task. Beyond model evaluation, \textsc{LLMAdBench} provides quantitative evidence on the advertiser-user trade-off and shows that the sponsorship disclosure systematically changes users' preference over ad placement.  
Together, these results position \textsc{LLMAdBench} as a public testbed for evaluating, diagnosing, and improving LLM systems that integrate advertising into generated content, while providing an empirical foundation for understanding how users respond to ads embedded in LLM outputs. 
\ifdefstring{\compilemode}{iclr}{Code and datasets are available at \url{https://anonymous.4open.science/r/LLMADBENCH/}.}{
Our datasets are available at \url{https://huggingface.co/datasets/ruiai-work/LLMADBENCH}.}
\end{abstract}
\section{Introduction}\label{sec:intro}

Large language models (LLMs) are rapidly becoming a new interface through which users make everyday decisions. Users nowadays increasingly ask LLMs to compare products, plan trips, choose services, and synthesize information before making purchases. This shift creates a natural opportunity for advertising, but also raises a new design challenge: \textbf{how can ads be incorporated without degrading the usefulness of underlying response?} 
One appealing approach is to leave the generated answer unchanged and introduce advertising only as an additional matched component ~\citep{openai2026testingads,perplexity2026advertising}. Such an approach separates monetization from the information used to support the user's decision and reduces the potential impact on consumer welfare. However, even when the underlying answer is held fixed, important design choices remain. Given the same user request, the same answer, and the same advertisement, where should the ad appear, and should it be explicitly disclosed? These seemingly simple choices motivate the central question of this work.

These choices matter because placement and disclosure can affect users and advertisers in different, and potentially conflicting, ways. An advertisement that appears at the beginning of a response may be easy to notice, but it may also delay the answer the user came for. An advertisement placed inside the answer may read more smoothly when it is unlabeled, yet feel intrusive once it is marked as sponsored. An advertisement at the end may be less disruptive, but it may also receive less attention. These are not small formatting choices. They determine how user experience, advertiser value, and transparency are traded against each another.

Existing datasets and evaluations have covered some aspects of advertising in LLM-generated content, but no single benchmark brings together the ingredients needed to study placement as a controlled behavioral question. As summarized in \Cref{tab:resource-comparison}, ad location is often chosen as part of a free-form generation or insertion process rather than isolated as an experimental variable. Human preference data are also limited, and disclosure effects and the reliability of LLM judges have received comparatively little systematic evaluation. This makes it difficult to determine whether a preference difference is driven by ad placement itself or by other changes in the generated content.


\begin{table}[t]
\centering
\small
\setlength{\tabcolsep}{1.5pt}
\renewcommand{\arraystretch}{1.05}
\caption{Comparison with related resources. ``Free-form'' means the ad location is chosen during generation or insertion rather than isolated as a controlled variable. ``Calib.'' denotes human-calibrated labels. ``Partial'' means the resource touches the dimension but does not evaluate it as a benchmark target.}
\label{tab:resource-comparison}
\resizebox{\linewidth}{!}{
\begin{tabular}{lcccccc}
\toprule
Resource & LLM ads & Human pref. & Controlled placement & Disclosure & Multi-objective & Judge reliability \\
\midrule
Webis Native Ads 2024 \citep{schmidt2024detecting} & $\checkmark$ & $\times$ & $\times$ & $\times$ & $\times$ & $\times$ \\
GEM-Bench \citep{hu2026gen} & $\checkmark$ & $\times$ & free-form & $\times$ & $\checkmark$ & $\times$ \\
NaiAD \citep{zhang2026naiad} & $\checkmark$ & calib. & free-form & $\times$ & $\checkmark$ & $\times$ \\
Click-intent evaluator \citep{turner2026evaluating} & $\checkmark$ & only 500 annotations & free-form & $\times$ & $\checkmark$ & partial \\
Ads that Talk Back \citep{tang2025ads} & $\checkmark$ & $\checkmark$ & $\times$ & partial & partial & $\times$ \\
LERA \citep{sun2026lera} & $\checkmark$ & $\times$ & free-form & $\times$ & partial & $\times$ \\
LLM-Auction \citep{zhao2025llm} & $\checkmark$ & $\times$ & free-form & $\times$ & $\checkmark$ & $\times$ \\
E-GEO \citep{bagga2025geo} & related & $\times$ & $\times$ & $\times$ & partial & $\times$ \\
\textsc{LLMAdBench} & $\checkmark$ & $\checkmark$ & $\checkmark$ & $\checkmark$ & $\checkmark$ & $\checkmark$ \\
\bottomrule
\end{tabular}
}
\end{table}

To address this gap, we introduce \textsc{\textbf{LLMAdBench}}, \textit{a human-preference benchmark that isolates ad placement while holding the underlying content fixed}. Each benchmark instance begins with a user query, an LLM-generated answer, and a matched advertisement, all of which are held fixed across comparisons. We construct four candidate responses that differ only in where the advertisement appears, say before the answer, early in the answer, late in the answer, or after the answer. We evaluate all pairwise placement comparisons under two disclosure conditions. In the unlabeled condition, the advertisement is inserted without an explicit marker. In the labeled condition, it is marked as \texttt{[Sponsored: \textit{$\langle$ad text$\rangle$}]}. Human annotators compare two candidates at a time and choose \emph{Version A}, \emph{Version B}, or \emph{Tie} along six dimensions covering \emph{naturalness}, \emph{visibility}, \emph{helpfulness}, \emph{click likelihood}, \emph{appropriateness}, and \emph{overall preference}. The benchmark contains 258 human-evaluated source conversations and 18,576 human judgments.

\subsection{Contributions}


This paper makes three contributions to the evaluation of advertising in LLM-generated content.

\paragraph{A controlled human-preference benchmark.}
We introduce \textsc{LLMAdBench}, a large-scale controlled human-preference benchmark that isolates the effect of ad placement in LLM responses. Each comparison fixes the user query, base answer, advertisement, and disclosure condition, and varies only the ad placement. This controlled design turns an otherwise messy interface choice into a clean pairwise evaluation task with A/B/T labels across six dimensions.

\paragraph{A systematic evaluation of LLM judges.}
We evaluate eight frontier LLM judges under the same protocol used for human annotators, and stress-test their judgments by swapping the presentation order of two candidates. Current frontier judges prove unreliable substitutes for human evaluation. Even the most stable models reverse roughly one quarter of their decisions under a swap, agreement across models is low, and their inferred placement preferences differ systematically from those of humans. \textsc{LLMAdBench} therefore provides a testbed not only for understanding human preferences, but also for evaluating the robustness, consistency, and human alignment of automated judges.


\paragraph{Empirical insights on human preferences.}
\textsc{LLMAdBench} provides new empirical evidence on how users respond to advertisements embedded in LLM-generated content. First, human preferences exhibit substantial learnable structure. A Qwen3-8B model fine-tuned on the human preference data substantially improves over its base model and outperforms all eight zero-shot frontier judges on the held-out prediction task. Second, human judgments reveal a systematic trade-off between advertiser value and user experience. Placements that make advertisements more visible tend to be less preferred overall, while explicit sponsorship disclosure shifts the preferred placement toward the end of the response. More broadly, the preferred insertion position depends on whether the advertisement is disclosed, revealing an important interaction between placement and disclosure. These findings provide potentially actionable guidance for the design of advertising in LLM systems.


\subsection{Related Work}
Webis Generated Native Ads studies detection of generated advertisements in conversational search responses \citep{schmidt2024detecting}. GEM-Bench and NaiAD provide broader resources for ad-injected response generation and evaluation \citep{hu2026gen,zhang2026naiad}. Other work studies click-intent evaluation, auction mechanisms, generative engine optimization, and user perceptions of advertising chatbots~\citep{turner2026evaluating,hajiaghayi2406ad,sun2026lera,bagga2025geo,tang2025ads}. These papers motivate the setting, but they do not provide a human-preference benchmark that cleanly varies placement and disclosure while keeping the underlying answer and chosen advertisement fixed. 
\ifdefstring{\compilemode}{iclr}{Due to the space limit, we discuss these and
more related works thoroughly in \Cref{app:related}.}{We discuss these and
more related works thoroughly in \Cref{app:related}.
}

\section{Task and Dataset}\label{sec:task-dataset}

\subsection{Task Definition}\label{subsec:task}

\textsc{LLMAdBench} evaluates where advertising content should appear in an LLM response. Each source instance begins with a user conversation/query $q$, a base response $r$, and a matched ad brief $a$. The benchmark keeps the conversation and the substantive answer fixed, then compares alternative ways of placing the advertisement around or inside the response. This setting reflects a practical constraint for user-facing LLM services that advertising is added to an existing answer, while the answer itself should remain the same to the user.

\begin{figure}[t]
\centering







\usetikzlibrary{arrows.meta,positioning}

\begin{tikzpicture}[
    font=\small,
    box/.style={
        draw=black!65,
        line width=0.5pt,
        rounded corners=2pt,
        text width=2.4cm,
        minimum height=1.15cm,
        align=center,
        inner sep=5pt
    },
    widebox/.style={
        box,
        text width=3.0cm
    },
    arrow/.style={
        -{Latex[length=2.0mm,width=1.3mm]},
        line width=0.5pt,
        draw=black!65
    },
    node distance=0.5cm
]

\node[box] (source) {
    conversation $q$\\
    base answer $r$\\
    ad brief $a$
};

\node[widebox, right=of source] (positions) {
    select four insertion\\
    positions with a\\
    fixed policy
};

\node[widebox, right=of positions] (pairs) {
    form all placement\\
    pairs under each\\
    condition
};

\node[box, right=of pairs] (labels) {
    collect A/B/T\\
    preferences\\
    on Q1-Q6
};

\draw[arrow] (source.east) -- (positions.west);
\draw[arrow] (positions.east) -- (pairs.west);
\draw[arrow] (pairs.east) -- (labels.west);

\end{tikzpicture}
\caption{\textsc{LLMAdBench} construction pipeline. Each source instance is expanded into four placement candidates and all pairwise placement comparisons. Candidate rendering follows a fixed insertion policy. Position rules, rendering prompts, and disclosure conditions are reported in \Cref{app:task-dataset}.}
\label{fig:task-pipeline}
\end{figure}
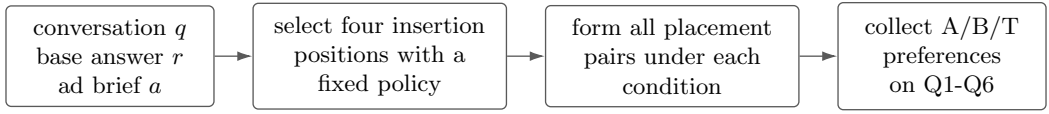

For each source instance, we construct four candidate responses at predefined insertion positions, say before the response, early in the response, late in the response, and after the response. We refer to these positions as \textsc{front}, \textsc{early}, \textsc{late}, and \textsc{end}. The front and end positions are fixed at the response boundaries, while the early and late positions are the two highest-scoring interior insertion points selected by the matching pipeline. Across the four candidates, we keep the conversation, the base answer, the advertised entity, and the core ad message fixed. The insertion policy may make local wording adjustments to the advertisement so that it reads naturally at its assigned position. The resulting comparisons measure how readers respond to placement under a fixed insertion procedure in \Cref{fig:task-pipeline}.
Each candidate also appears under two disclosure conditions. In the \emph{unlabeled} condition, the inserted advertisement has no explicit sponsorship marker. In the \emph{labeled} condition, the inserted advertisement is marked with \texttt{[Sponsored: \textit{$\langle$ad text$\rangle$}]}. The label changes only the disclosure status of the inserted content. The placement comparison remains within the same source instance and insertion pipeline.

The evaluation unit is a pairwise comparison between two placements. For a fixed source instance and disclosure condition, the four placements form a complete tournament with $\binom{4}{2}=6$ unordered placement pairs. Annotators see two candidate responses as Version A and Version B, with presentation order randomized, and answer the same preference questions for each pair. They may choose Version A, Version B, or Tie. We keep ties as labels rather than discarding them, since indifference is meaningful when two placements differ only in where the same ad brief is realized.

\subsection{Preference Rubric}\label{subsec:rubric}

Ad placement can affect multiple objectives simultaneously, so a single quality score is insufficient to characterize its impact.  A front placement may make the advertisement easy to notice, but it can also interrupt the answer. A later placement may preserve the response flow at the cost of visibility. Disclosure may further change how the same position is perceived. \textsc{LLMAdBench} therefore collects preferences along six dimensions under two disclosure conditions, shown in \Cref{tab:rubric}.

The rubric is designed to capture the tension between user experience and advertiser value. Naturalness of flow, content helpfulness, and appropriateness measure how the insertion affects the user's experience.  Ad visibility and click likelihood capture advertiser-relevant outcomes. Overall preference gives the all-things-considered assessment. Importantly, visibility is defined as a neutral perceptual judgment. Greater visibility does not by itself imply a better placement. For click likelihood, we ask annotators to report which advertisement they would be more likely to click if required to choose one, providing a measure of stated behavioral intention rather than observed clicking behavior.


\begin{table*}[t]
\centering
\small
\setlength{\tabcolsep}{5.5pt}
\renewcommand{\arraystretch}{1.08}
\begin{tabular}{llp{10.2cm}}
\toprule
& Dimension & Definition given to annotators \\
\midrule
Q1 & Naturalness of flow &
Which version reads more smoothly and feels less disruptive. A natural insertion fits the surrounding response rather than appearing abruptly. \\
Q2 & Ad visibility &
In which version is the advertisement easier to notice. This is a neutral perceptual judgment that more visible is neither better nor worse by itself. \\
Q3 & Content helpfulness &
Which version better preserves the usefulness of the non-ad content for answering the user's question, irrespective of the ad's own usefulness. \\
Q4 & Click likelihood &
If forced to click one advertisement, which version's ad the annotator would be more likely to click. \\
Q5 & Appropriateness &
Which version feels more trustworthy and less manipulative, including whether the ad is suitable, transparent, and not overly pushy. \\
Q6 & Overall preference &
Which version the annotator prefers, all things considered. \\
\bottomrule
\end{tabular}
\caption{Preference rubric used for human annotation. Each question is answered as Version A, Version B, or Tie for the same pair of candidate placements.}
\label{tab:rubric}
\end{table*}

\subsection{Dataset Construction}\label{subsec:dataset-construction}

The benchmark is built from 258 human-evaluated source conversations. Each source conversation is paired with one matched ad brief and expanded into four candidate placements under both disclosure conditions. This yields 1,548 pairwise comparisons per condition and 18,576 dimension-level human judgments in total. The unlabeled and labeled conditions use the same conversations, base answers, ad briefs, and placement pairs. This paired design isolates the effect of the disclosure marker, and we include more details in \Cref{app:task-dataset}. 

The source conversations include both single-turn requests and multi-turn contexts. When prior turns are available, they are shown to annotators together with the current request. The inserted advertisements are short relative to the base responses, with a median advertisement length of 25 words and a median response length of 194 words in the human-evaluated subset. The resulting task is therefore native placement within an existing answer, not a choice between an answer and a separate advertising block.

\subsection{Human Annotation}\label{subsec:human-annotation}
Human annotators evaluate placement pairs using the A/B/T interface described above. Each task shows the conversation context and two candidate responses, with candidate order randomized across comparisons. We use the same interface in the unlabeled and labeled conditions, so differences between conditions are estimated from the same placement comparisons with and without the label.

The main evaluation uses labels that pass our quality-control filters, including an attention-check question with a verifiable answer. To test whether the benchmark contains a stable aggregate human signal, we repeatedly split the human-annotated data into random halves and re-estimate placement preferences in each half. Split-half reliability is high for overall preference in both conditions with 0.975 in the labeled condition and 0.966 in the unlabeled condition. Note that these estimates do not define a human performance ceiling. They show that the aggregate preference signal used for model-judge evaluation is reproducible under resampling. Full reliability results are reported in \Cref{subsec:human-signal-stability}.

\section{Evaluation Protocol}\label{sec:evaluation-protocol}

\subsection{Prediction Task}\label{subsec:prediction-task}

A judge is given the task-relevant materials shown to annotators, including the user conversation, two rendered candidate responses, the disclosure condition, and one target dimension from the rubric. It predicts the retained human annotation for that dimension in the three-class label space \{Version A, Version B, Tie\}, after the quality-control filter described in \Cref{app:annotation-qc}. The auxiliary placement-attention item shown in the human interface is used only for quality control and will not be a benchmark target.

We report exact three-way accuracy separately for the labeled and unlabeled conditions. Ties are kept as a separate class because they mark cases in which readers did not prefer either placement on the specified dimension. For the fine-tuned judges, splits are made at the source-instance level, so comparisons derived from the same conversation and advertisement cannot appear on both sides of a train-test split. Zero-shot frontier judges use the same A/B/T label space and prompt format detailed in \Cref{app:prediction-details,app:model-judge-prompt}.

\subsection{Placement Preference Estimation}\label{subsec:placement-estimation}

The A/B/T labels can also be used to estimate which of the four positions is preferred on average. For each disclosure condition and target dimension, we fit a Bradley-Terry model \citep{bradley1952rank} with Davidson's tie extension \citep{davidson1970ties} and a presentation-order term \citep{davidson1977order} as 
\begin{align*}
\Pr(A \succ B) &=
\frac{v_A e^\delta}
{v_A e^\delta + v_B + \nu \sqrt{v_A v_B e^\delta}}, \\
\Pr(B \succ A) &=
\frac{v_B}
{v_A e^\delta + v_B + \nu \sqrt{v_A v_B e^\delta}}, \\
\Pr(\mathrm{tie}) &=
\frac{\nu \sqrt{v_A v_B e^\delta}}
{v_A e^\delta + v_B + \nu \sqrt{v_A v_B e^\delta}},
\end{align*}
where $v_A$ (resp., $v_B$) is the position value (strength) associated with the insertion position of Version A (resp., Version B), $\nu\ge 0$ controls the propensity to declare a tie and $\delta$ captures any systematic presentation-order effect.
The four position strengths are normalized to sum to one and reported as placement scores. For example, a click-likelihood placement score is the normalized position strength estimated from Q4 labels. 
Under no systematic preference among positions, each score is close to $0.25$, so scores above $0.25$ indicate positions that are favored more often under the fitted pairwise-choice model. The full likelihood and the split-half reliability test for these placement scores are reported in \Cref{app:human-reliability}.

We report placement scores in addition to three-way prediction accuracy because the two metrics test different forms of agreement with humans. Prediction accuracy asks whether a judge matches the human A/B/T label on each comparison. Placement scores ask whether the judge recovers the same aggregate ordering and relative strength of \textsc{front}, \textsc{early}, \textsc{late}, and \textsc{end}. A judge can do reasonably well on individual comparisons while still imposing a systematically different placement preference, so both views are needed for a benchmark of ad placement.

\subsection{Reliability Diagnostics}\label{subsec:reliability-diagnostics}

We use three diagnostics to characterize the reliability of human and model judgments from complementary perspectives. First, each model judge is evaluated under a presentation-swap test. The two candidate responses are swapped between Version A and Version B while their text remains unchanged. For non-tie predictions, consistency requires the answer to flip from Version A to Version B, or from Version B to Version A. For ties, consistency requires Tie to remain Tie. Swap error is the fraction of comparisons for which the judge fails to map A to B, B to A, or Tie to Tie after the two candidate responses are exchanged, and we discuss more in \Cref{app:swap-error}.


Second, we measure agreement across different language model judges. Raw agreement can be misleading when judges have similar label-use patterns. For example, when both judges frequently predict Tie, the agreement can be inflated even if their item-level judgments are not strongly aligned.  We therefore use chance-corrected agreement where pairwise agreement is computed for each pair of models, and panel-level agreement is computed across the full model-judge panel. Both diagnostics use the same A/B/T label space as the benchmark, with more details reported in \Cref{app:judge-agreement}.

Third, we report split-half reliability for the human aggregate placement signal. The split is performed over source instances, and the placement model is re-estimated on each half. Reliability is computed from the agreement between the two normalized placement-score vectors, with the full procedure reported in \Cref{app:human-reliability}. Together, the protocol distinguishes disagreement with humans, presentation-order sensitivity, disagreement among judges, and stability in aggregate placement estimates.
\section{Benchmark Results}\label{sec:benchmark-results}

\subsection{Models and Baselines}\label{subsec:models-baselines}

We evaluate \textsc{LLMAdBench} as a judge benchmark. The main comparison includes eight zero-shot frontier LLM judges including GPT-5.4~\citep{singh2025openai}, Claude-Opus-4.6~\citep{anthropic2026claudeopus46}, Gemini-3.1-Pro~\citep{gemini31pro2026}, Grok-4.1~\citep{grok41_model}, Kimi-K3~\citep{team2026kimi}, DeepSeek-V4-Pro~\citep{xu2026deepseek}, Llama-4-Maverick~\citep{meta2025llama4}, and Mistral-Large-3~\citep{mistrallarge32025}. Each judge receives the same pairwise A/B/T prompt and predicts the retained human label for each single rubric dimension.

We also include Qwen3-8B~\citep{yang2025qwen3} before and after preference training on \textsc{LLMAdBench}. The adapted model uses Direct Preference Optimization~\citep{rafailov2023direct} on human comparisons from the training split, with evaluation performed on the held-out instances the same as for the eight frontier LLMs. This gives two different reference points, namely zero-shot general-purpose judges and a task-adapted judge trained from human preferences over ad placement.



\subsection{Main Leaderboard}\label{subsec:main-leaderboard}

\Cref{tab:judge-accuracy} reports held-out A/B/T prediction accuracy. We treat each placement-pair–rubric combination as a separate prediction task. 
The zero-shot frontier judges occupy a tight range, with the best score being 48.14 in the labeled condition and 50.32 in the unlabeled condition. Qwen3-8B starts below this range before preference training, but reaches 50.54 and 55.76 after DPO. The adapted judge therefore improves over its own base model by 10.29 points in the labeled condition and 19.24 points in the unlabeled condition, while also exceeding the strongest zero-shot frontier judge by 2.40 and 5.44 points. Paired McNemar tests~\citep{mcnemar1947note} with Holm correction~\citep{holm1979simple} confirm that the post-DPO judge improves over every frontier judge in both conditions, with adjusted $p$-values below $0.0088$ for labeled comparisons and below $2.7\times10^{-7}$ for unlabeled comparisons. Full test statistics are reported in \Cref{app:benchmark-paired-tests}. 



\begin{table*}[!bh]
\centering
\small
\setlength{\tabcolsep}{4.2pt}
\renewcommand{\arraystretch}{1.08}
\begin{tabular}{lcccccccccc}
\toprule
Condition
& GPT-5.4& Claude& Gemini& Grok& Kimi& DeepSeek& Llama-4& Mistral& Pre-DPO& Post-DPO \\
\midrule
Labeled& 47.85& 46.96& 48.14& 46.31& 47.23& 44.00& 44.51& 46.39& 40.25& \textbf{50.54} \\
Unlabeled& 50.32& 48.74& 49.92& 47.52& 49.87& 46.18& 48.09& 48.76& 36.52& \textbf{55.76} \\
\bottomrule
\end{tabular}
\caption{Held-out A/B/T prediction accuracy. The first eight columns are zero-shot frontier judges, and the last two columns show Qwen3-8B before and after five DPO epochs.}
\label{tab:judge-accuracy}
\end{table*}

\begin{figure}[t]
\centering
\includegraphics[width=\linewidth]{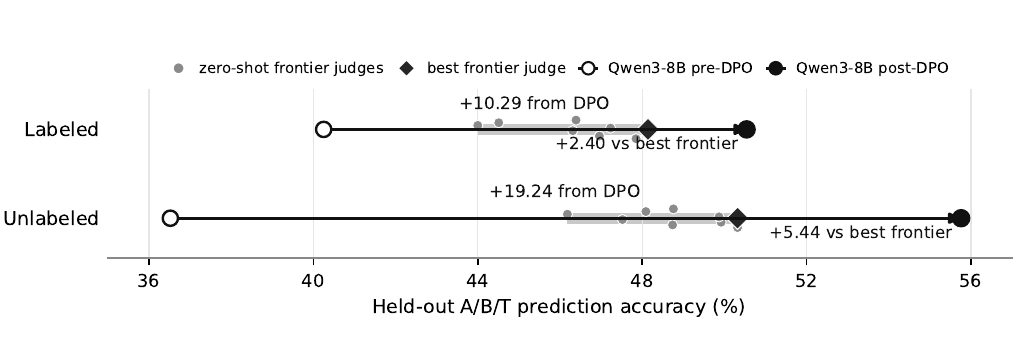}
\caption{Zero-shot reference range and DPO shift on held-out A/B/T prediction. Gray points are frontier LLM judges under the shared zero-shot prompt, and black markers show Qwen3-8B before and after DPO.}
\label{fig:dpo-reference-range}
\end{figure}

\Cref{fig:dpo-reference-range} summarizes the same result as a reference-range comparison. The gray points show the eight zero-shot frontier judges, the diamond marks the best zero-shot judge in each condition, and the black shift shows Qwen3-8B before and after five DPO epochs. The figure shows that the benchmark contains a preference signal that can be learned from human comparisons, but is not recovered by zero-shot prompting alone.

\subsection{What Preference Training Adds}\label{subsec:dpo-gain}

The main result is not that an 8B base model is a stronger general judge than frontier LLMs. Before DPO, Qwen3-8B is the weakest judge in both conditions. The shift after DPO is a useful signal that supervision from \textsc{LLMAdBench} moves the same model from below the zero-shot frontier range to significantly above it. This suggests that the benchmark captures placement preferences that are learnable from human comparisons, but are not reliably recovered by prompting strong general-purpose judges. The effect is larger in the unlabeled condition, where the model cannot rely on an explicit disclosure marker and must infer how placement changes the reading experience itself.


\section{Why Human Evaluation Is Needed}
\label{sec:judge-reliability}

A benchmark for ad placement needs a reproducible human-preference target. Since the inputs are text-only and the rubric is explicit, one plausible shortcut is to replace human labels with zero-shot LLM judgments~\citep{zheng2023judging}. We evaluate that shortcut with two minimum checks for automatic annotators, including invariance to a semantics-preserving A/B presentation swap, and chance-corrected agreement with other judges on the same evaluation units. \textsc{LLMAdBench} separates these roles where human judgments define the target signal, while LLM judges are analyzed as automatic evaluators rather than used to define labels.

\subsection{Presentation-Swap Instability}
\label{subsec:swap-instability}

The presentation-swap test is an invariance check within a single judge. For each LLM, we run the full evaluation twice, once with the original Version A/B order and once with the two candidate responses exchanged. A stable judge should reverse A and B decisions after the swap and preserve Tie decisions. 

\Cref{fig:judge-reliability-summary} shows that every zero-shot judge is sensitive to presentation order. Across all six rubric dimensions, the lowest average swap error is $24.36\%$, and the eight-judge mean is $31.40\%$. The instability is present in both disclosure conditions, with mean swap error $28.93\%$ in the labeled condition and $33.87\%$ in the unlabeled condition. It exhibits non-negligible inconsistencies within the same model on two presentations of the same comparison.

\subsection{Low Agreement Among LLM Judges}
\label{subsec:judge-agreement-main}

\begin{figure}[tbp]
    \centering
        \begin{minipage}{0.52\textwidth}
        \centering
        \includegraphics[width=\linewidth]{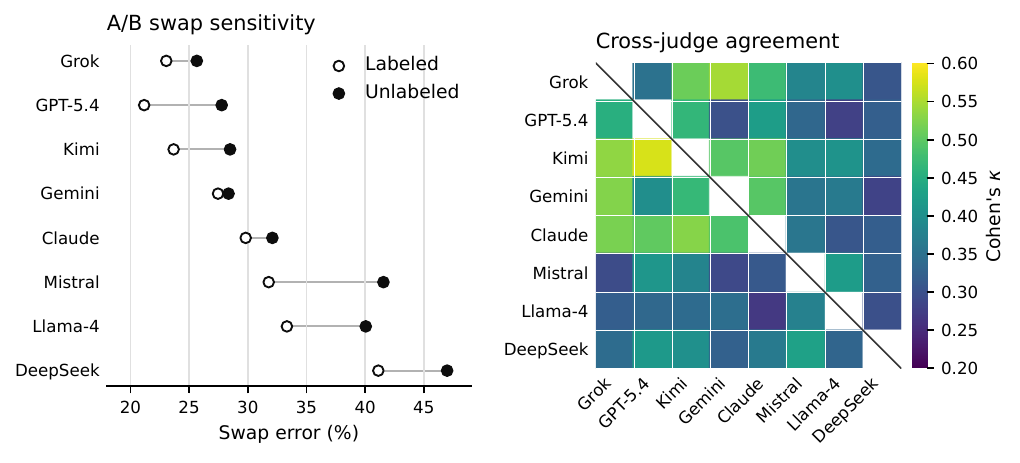}
        \vspace{-18pt}
\caption{Left: Reliability diagnostics for eight zero-shot LLM judges. Right: The Pairwise Cohen's $\kappa$ in the labeled condition with explicit ad disclosure (lower triangle) and the unlabeled condition without disclosure (upper triangle).}
\label{fig:judge-reliability-summary}
    \end{minipage}
    \hfill
    \begin{minipage}{0.47\textwidth}
        \centering
 \includegraphics[width=\linewidth]{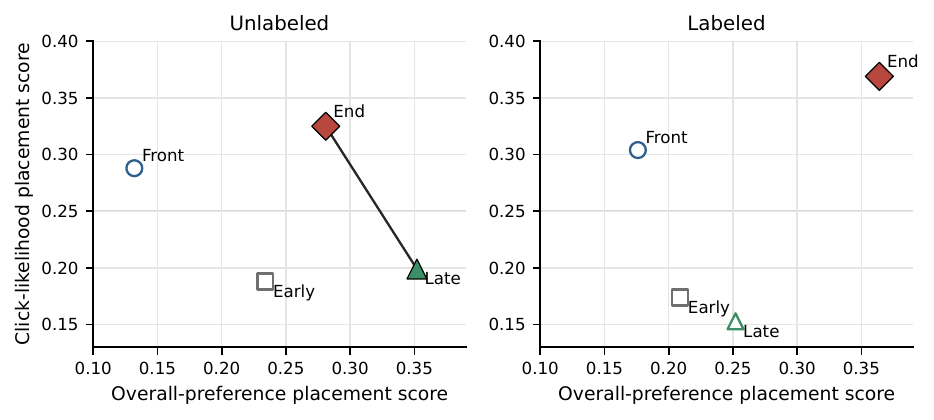}
\caption{Human Davidson-Bradley-Terry placement scores for overall preference and click likelihood. Filled markers indicate positions that are Pareto-efficient under the point estimates within each disclosure condition.}
\label{fig:human-pareto-frontier}
    \end{minipage}
\end{figure}

Swap consistency is an individual-judge requirement. We next ask how much different judges agree with one another. We use Cohen's $\kappa$ as a chance-corrected measure of agreement between pairs of model judges, with the full definition reported in \Cref{app:judge-agreement}. On the original A/B presentation, the mean pairwise $\kappa$ is $0.402$ in the labeled condition and $0.384$ in the unlabeled condition. The heatmap in \Cref{fig:judge-reliability-summary}, which computes pairwise Cohen's $\kappa$ between model judges on the original A/B presentation, shows that the limited agreement is not driven by a single outlier model. 

Combining multiple zero-shot judges does not close this performance gap. When the eight zero-shot judges are combined by majority voting on the same held-out A/B/T targets~\citep{wang2022self}, the ensemble remains significantly less accurate than the post-DPO Qwen3-8B judge in both disclosure conditions. Paired McNemar tests give a p-value of 0.0037 for labeled and $4.7\times10^{-8}$ for unlabeled, respectively. More detailed results are reported in \Cref{app:judge-agreement-full}.

Together, these diagnostics argue against using these zero-shot LLM judges as the sole source of benchmark labels. They are sensitive to response order despite unchanged response content, and different judges show only limited agreement on the same A/B/T decisions.

\subsection{Human Signal Stability}
\label{subsec:human-signal-stability}
We perform the split-half test at the source-instance level to test whether the aggregate human placement signal is reproducible under resampling. In each split, we re-estimate the Davidson-Bradley-Terry placement model on two halves of the human-annotated data and compare the resulting position-preference vectors. 
This split-half test measures whether the aggregate placement preference remains stable when estimated from independent halves of the human-annotated data.
Full details are reported in \Cref{app:human-aggregate-reliability-full}.

By this aggregate split-half criterion, the human signal is stable. Split-half reliability ranges from $0.975$ to $0.998$ across the six dimensions in the labeled condition and from $0.952$ to $0.999$  in the unlabeled condition. Overall preference has reliability of $0.975$ and $0.966$, respectively. These results support using aggregated human judgments as the benchmark target. 
\section{What \textsc{LLMAdBench} Reveals}
\label{sec:benchmark-findings}

\textsc{LLMAdBench} also provides an empirical platform for studying how users respond to ad-inserted LLM content.
We summarize these patterns with the Davidson-Bradley-Terry placement scores defined in \Cref{sec:evaluation-protocol}. Since the four position scores are normalized to sum to one, $0.25$ is the equal-share reference point and values above it indicate above-baseline estimated preference mass.

\subsection{Disclosure Changes the Preferred Placement}
\label{subsec:disclosure-placement}

Human preferences over insertion positions depend on whether the advertisement is disclosed. Front placement is disfavored in both conditions, with an overall-preference score of $0.176$ with disclosure and $0.132$ without disclosure. The preferred non-front position changes with disclosure. In the unlabeled condition, the late in-response position has the highest overall-preference score, $0.352$ with 95\% CI $[0.317,0.389]$, while end placement is lower at $0.281$ with CI $[0.252,0.313]$. In the labeled condition, end placement becomes the clear preference, increasing to $0.364$ with CI $[0.329,0.399]$, while late placement falls to $0.252$ with CI $[0.223,0.287]$. \textsc{LLMAdBench} reveals a reversal in human preferences between late and end placement across disclosure conditions. This highlights the importance of using disclosure-condition-specific human benchmarks for evaluation.


\subsection{Overall Preference and Click Likelihood Diverge}
\label{subsec:pareto-tradeoff}

\Cref{fig:human-pareto-frontier} plots each position by two human placement scores, with overall preference on the horizontal axis and click likelihood on the vertical axis. A position is Pareto-efficient under these point estimates when no other position has both a higher overall-preference score and a higher click-likelihood score.


Using the point estimates, the unlabeled condition shows a trade-off. Late placement achieves the highest overall preference ($0.352$) but a lower click-likelihood score ($0.199$), whereas end placement has lower overall preference ($0.281$) but the highest click likelihood ($0.325$). Front and early placements are Pareto dominated.
Under disclosure, this trade-off disappears. End placement has the highest point estimate for both overall preference, $0.364$, and click likelihood, $0.369$. The two disclosure conditions therefore induce different placement frontiers under the human scores.

\subsection{LLM Judges Collapse Toward End Placement}
\label{subsec:llm-placement-pattern}
The same placement-score analysis reveals a systematic gap between human readers and zero-shot LLM judges. For overall preference, human judgments distinguish late and end placement depending on disclosure. The frontier LLM judges are much less conditional. With one exception, Mistral-Large-3 in the unlabeled condition, every other frontier judge assigns its highest overall-preference score to end placement in both disclosure conditions. Several judges make this preference highly concentrated. Claude-Opus-4.6 assigns $0.688$ of its unlabeled overall-preference mass to end placement and $0.846$ in the labeled condition, compared with human scores of $0.281$ and $0.364$.

This placement-score discrepancy is consistent with the reliability results in \Cref{sec:judge-reliability} that zero-shot judges do not simply reproduce the human target. The discrepancy is visible not only in individual A/B/T predictions, but also after aggregating choices into placement scores. Full placement-score tables for humans and LLM judges are reported in \Cref{app:human-placement-results}.
\section{Conclusion}

\textsc{LLMAdBench} frames ad placement in LLM responses as a controlled human-preference benchmark. By fixing the query, base answer, matched advertisement, disclosure condition, and candidate insertion positions, it makes the placement decision measurable without folding it into general response quality. The results show why this control is useful. Eight zero-shot frontier judges are unstable under A/B presentation swaps, agree only moderately with one another, and often collapse toward end placement rather than the disclosure-dependent human pattern. With direct supervision, a small Qwen3-8B judge trained on the collected comparisons outperforms the zero-shot judges on held-out human-label prediction. The human labels also reveal structure that a single ad-quality score would hide, such as front placement receives high visibility scores but is penalized overall, disclosure moves the preferred non-front placement from late in the response to the end, and click likelihood follows a different profile from overall preference.

We establish \textsc{LLMAdBench} as a testbed for evaluating ad-insertion policies and model judges under the same human-preference target. Future work can extend the benchmark to richer ad formats, longer interaction histories, additional disclosure interfaces, and submitted judge models. The broader goal is to make ad-supported LLM interfaces measurable enough to study trade-offs among user preference, click-oriented outcomes, and platform design~\citep{feizi2025online}.

\bibliography{iclr2027_conference}
\bibliographystyle{iclr2027_conference}

\appendix
\crefalias{section}{appendix}
\crefalias{subsection}{appendix}
\section{Omitted Details in \Cref{sec:intro}}
\subsection{Related Work}\label{app:related}

\citet{feizi2025online} provides an early conceptual framework that separates LLM advertising into four modules, including modification, bidding, prediction, and auction. Subsequent research, however, does not map one-to-one onto these modules, where bidding is generally studied together with allocation and pricing, while ad insertion, detection, and user perception have developed into distinct research questions. We therefore organize the related literature into three areas.

\paragraph{Auction mechanisms for LLM advertising.}
Existing auction mechanisms for LLM advertising differ mainly in what advertisers bid for and how their bids influence the resulting content. \citet{duetting2024mechanism} formulates a token-level auction in which advertiser-specific LLMs influence generation through scalar bids. \citet{soumalias2026truthful} instead aggregates advertisers' preferences over the complete response.  \citet{hajiaghayi2406ad} introduces RAG-based segment auctions, which can retrieve ads according to relevance and bids. These approaches reveal a common trade-off. Token-level and response-level mechanisms can incorporate rich contextual information but often require repeated LLM inference, whereas retrieval-based mechanisms are more efficient but may rely on coarse relevance signals. Subsequent work explores different ways to preserve allocation quality while reducing the computational costs.
\citet{zhao2025llm} integrates ad allocation and response generation within a learned mechanism, allowing the model to capture interactions among ads without separate inference for every advertiser. By contrast, \citet{xu2026ad} deliberately decouples ad insertion from response generation and decouples bidding from individual response contexts. \citet{sun2026lera} further improves retrieval-based allocation through embedding filtering followed by LLM-logit reranking, increasing ad-selection accuracy. 

\paragraph{Ad generation and insertion.}
A second stream of work studies how sponsored content is produced once the advertising objective or selected ad is given. \citet{reisenbichler2026applying} tailors an open-source LLM to generate sponsored-search ad copy designed to attract clicks. \citet{liu2025llms} develops MarketingFM, a retrieval-augmented system that generates keyword-specific ad copy for offsite search marketing. Focusing on persuasive effectiveness,
\citet{meguellati2025llm} finds that LLM-generated ads perform comparably to human-written ads under personality-based personalization and outperform them when applying general persuasion principles. These studies focus primarily on standalone advertisements. Moving to advertisements embedded within LLM responses, \citet{zhang2026pila} proposes PILA, a model-agnostic sidecar that rewrites an existing response to integrate sponsored content without modifying the upstream model or workflow, which improves ad effectiveness while preserving response quality.

\paragraph{Evaluation and user perception.}
A third stream of work raises two related questions, namely how to evaluate an ad-embedded response and how users actually perceive it. \citet{zhang2026naiad} introduces NaiAD, which provides 58,999 ad-embedded responses with calibrated measures that separate user utility from commercial utility, enabling the above two objectives to be studied and controlled independently. \citet{turner2026evaluating} constructs simulated behavioral supervision, then distills it into a continuous evaluator of click-through intent and related dimensions of ad quality. In a 179-participant study, \citet{tang2025ads} finds that users often fail to recognize personalized ads embedded in chatbot responses and sometimes prefer undisclosed advertising responses. Participants also tend to manage advertising through conversational requests rather than a conventional disclosure interface. Together, these studies provide evaluators, benchmarks, and evidence about the general perception of embedded advertising, but they do not identify how the same ad is perceived at different response positions or how that relationship changes under explicit sponsorship disclosure. Our study isolates these effects through controlled human comparisons.

\section{Omitted Details in \Cref{sec:task-dataset}}\label{app:task-dataset}

\subsection{Data Sources and Matching}\label{app:data-sources}

\textsc{LLMAdBench} combines conversational contexts from WildChat~\citep{zhao2024wildchat} with advertisement material derived from the Webis Generated Native Ads 2024 dataset~\citep{heineking2025webis}. The processing unit is a single WildChat conversation turn. For retrieval, we concatenate the current user question with the corresponding assistant response. Earlier turns are not used to retrieve advertisements; they are recovered later only for survey presentation.

The Webis materials provide GPT-4-generated native-ad sentences paired with surrounding response text. We use the positive native-ad pairs as a source of ad stimuli, extracting the sentence that carries the advertised brand or product and its core message. The resulting pool supplies short, traceable advertisement units that can be matched to WildChat turns and rendered at different response positions while keeping the advertised entity fixed within each comparison.

The matching procedure consists of two steps. We use Qwen3-Embedding-0.6B~\citep{zhang2025qwen3} to encode the current user-assistant exchange and the advertisement candidates. Candidate advertisements here are ranked by cosine similarity. In practice, the ranking may also be affected by advertisers’ bids or other factors. For each conversation turn, the retrieval stage keeps up to five advertisements with the highest similarity scores, subject to a threshold of $0.65$. The threshold removes turns without a sufficiently relevant advertisement match.
Qwen3-8B then filters the retrieved candidates. The judge sees the dialogue, the candidate advertisement keyword or content, and the supporting advertisement sentence, and returns a single Boolean field, \texttt{relevant}. Candidates marked irrelevant are removed before placement construction. This filtering step is used only to construct candidate materials, and benchmark labels come from human pairwise preferences.

{\bf System Prompt}
\begin{Verbatim}[breaklines=true, breaksymbol=]
You are a relevance judge for ad insertion.

Given a dialogue and one candidate advertisement, determine whether the advertisement is relevant to the dialogue.

Set "relevant" to true only when the advertisement is relevant to the dialogue. Otherwise, set it to false.

Output rules:
- Return exactly one valid JSON object using the following schema: {"relevant": true}
- The value of "relevant" must be a JSON boolean: true or false.
- Do not include explanations, Markdown, code fences, or additional fields.
\end{Verbatim}

{\bf User Prompt}
\begin{Verbatim}[breaklines=true, breaksymbol=]
Dialogue: {conversation_text}
Candidate advertisement keyword/content: {advertisement}
Supporting advertisement sentence: {ad_sentence}

Is this advertisement relevant to the dialogue?
Return exactly one valid JSON object using the following schema: {"relevant": true}

Set "relevant" to true if the advertisement is relevant and false otherwise. Do not include any other text or fields.
\end{Verbatim}
After the automatic pipeline produces more than three thousand candidate records, human reviewers curate the pool before preference annotation. The curation removes conversations that were meaningless or lacked research value, non-English records, records with abnormal symbols or malformed formatting, incomplete text, overly specialized conversations that ordinary annotators could not assess reliably, and conversations requesting prompt writing, drawing commands, or similar specialized technical instructions. When more than one advertisement candidate remained for the same WildChat turn, the curated pool kept one candidate for that turn.

\subsection{Position Selection}\label{app:position-selection}

For each retained conversation-advertisement pair, we construct four placement candidates. Position 1 is the beginning of the assistant response, and Position 4 is the end. Positions 2 and 3 are selected from internal sentence-boundary insertion points.

Let $s_i$ and $s_{i+1}$ denote the adjacent sentences before and after an internal boundary $i$, and let $a$ denote the advertisement sentence. Using our embeddings, the pipeline scores each internal boundary by
\[
D_i =  \cos(s_i, a) + \cos(a, s_{i+1})-2\cos(s_i, s_{i+1}).
\]
This score compares the original local continuity with the continuity after inserting the advertisement between the two neighboring sentences. Higher scores indicate smaller local disruption. The pipeline selects the two internal boundaries with the highest scores and orders them by their occurrence in the original response. The earlier selected boundary defines Position 2, and the later selected boundary defines Position 3. We therefore use this procedure to choose the four candidate positions before human annotation.

\subsection{Candidate Rendering}\label{app:candidate-rendering}

The placement candidates are rendered with Qwen3-8B after the four positions have been selected. The model receives the original assistant response, the original advertisement sentence, and the local left and right context around the target insertion point. It is instructed to rewrite only the advertisement text so that it fits between the two local contexts, while preserving the core advertising information and the advertised product, brand, or entity name. 

{\bf System Prompt}
\begin{Verbatim}[breaklines=true, breaksymbol=]
You are an ad rewriter for local context insertion.

Task:
- Rewrite ONLY the ad text so it can be inserted between the given left and right context.

Rewrite requirements:
1. Make left context + rewritten ad + right context logically SMOOTH.
2. Keep the core ad information, but do NOT need to follow the original ad sentence wording.
3. You MUST preserve the ad/product/brand name from the original advertisement text.

Output rules:
- Return only rewritten ad text.
- No explanation, no labels, no quotes.
\end{Verbatim}
{\bf User Prompt}
\begin{Verbatim}[breaklines=true, breaksymbol=]
Original assistant response: {original_response}
Original advertisement: {matched_ad}
Left context around the insertion point: {left_context or "[EMPTY]"}
Right context around the insertion point: {right_context or "[EMPTY]"}

Task:
Rewrite the advertisement so it can be inserted smoothly between the left and right context.

Rewrite guidance:
- Rewrite ONLY the ad
- Change wording freely, but you need to preserve the core ad information
- You MUST keep the ad/product/brand name from the original advertisement

Return only the final rewritten text.
\end{Verbatim}
Rendering is performed separately at the four selected positions, allowing the advertisement to take the local wording needed for each context. The benchmark comparison remains controlled at the level that matters for placement, including the same conversation, base answer, advertised entity, and advertising message are shared across the paired candidates. Recovered multi-turn context is used only to assemble the materials shown to annotators; it is not used for advertisement retrieval, position selection, or rendering.

\subsection{Annotation Interface and Quality Control}\label{app:annotation-qc}
We recruit 12 students and researchers affiliated with universities in the United States, Canada, and China, all with strong proficiency in English. For each conversation-advertisement instance, the four rendered responses define six unordered placement pairs. These pairwise tasks are distributed across our participants. Within each participant assignment, task order is shuffled under the constraint that the same advertised entity does not appear in consecutive questions. The left-right order of each pair is also randomized.

\begin{figure}[!htp]
    \centering
    \includegraphics[width=\linewidth]{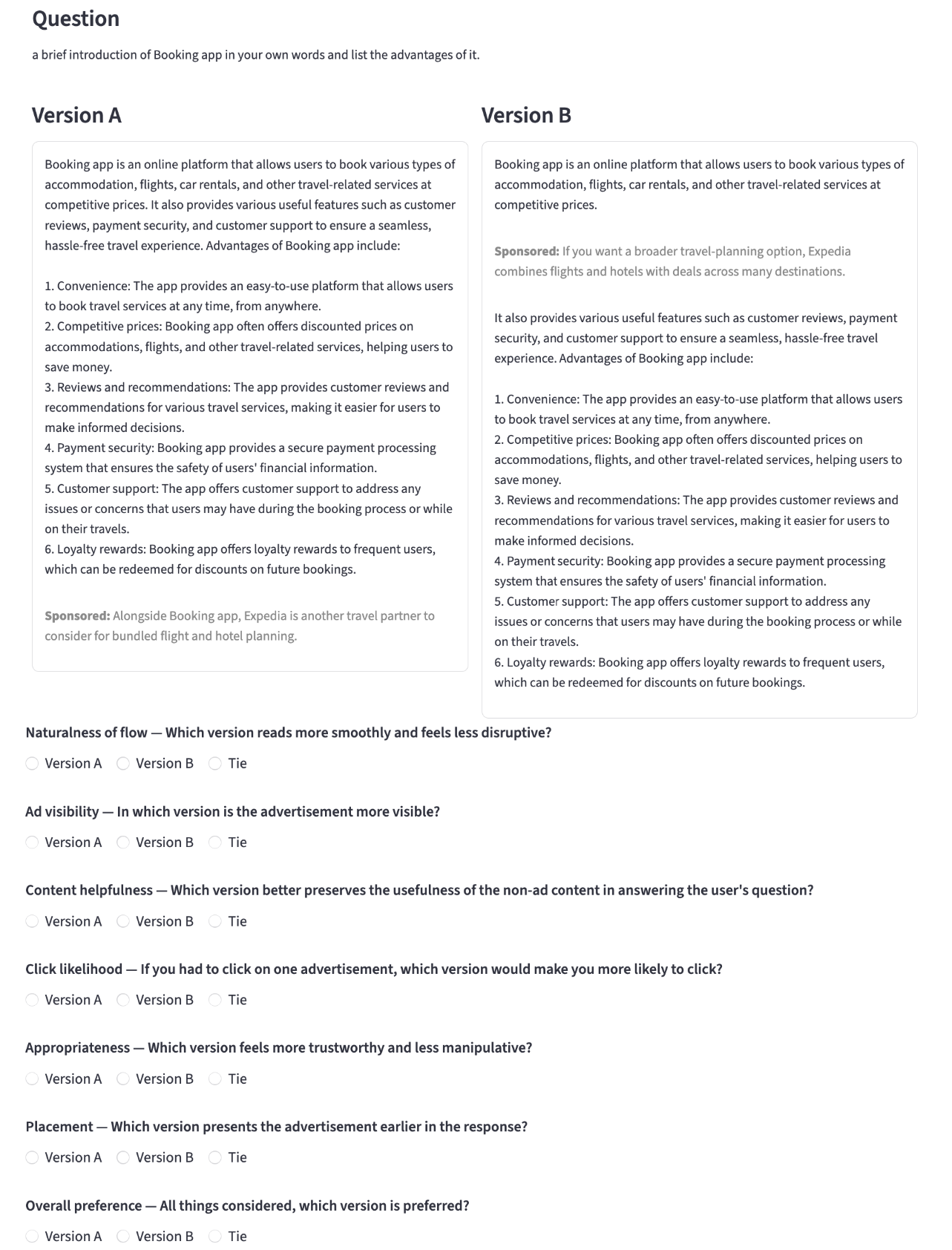}
    \caption{Human evaluation interface for pairwise placement judgments. Annotators compare two candidate responses for the same user query and answer the rubric questions with A/B/T.}
    \label{fig:annotation-interface}
\end{figure}

Annotators see the conversation context and two candidate responses. For each pair, they answer the six rubric questions in \Cref{tab:rubric} by choosing Version A, Version B, or Tie. \Cref{fig:annotation-interface} shows the interface used for pairwise candidate evaluation. The randomized presentation prevents a placement from being tied to a fixed screen side or a repeated local task order.

Quality control is performed through two attention checks with verifiable answers. The first asks which version places the advertisement earlier in the response. For records that pass this check, we compare the submitted advertisement text with the target sentence, requiring a Gestalt pattern-matching similarity ratio above $0.8$~\citep{enwiki:1305354093}. We further conduct a second annotation round to replace records that fail either check. All annotations collected in the second round pass both checks and replace the corresponding failed records.

\subsection{Aggregate Human Signal Reliability}\label{app:human-reliability}

We assess reliability at the level used by the benchmark, say aggregate preferences over insertion positions. For each disclosure condition and rubric dimension, we estimate a Davidson-Bradley-Terry choice model \citep{bradley1952rank,davidson1970ties,davidson1977order}. Let $v_k>0$ denote the latent preference value of position $k\in\{1,2,3,4\}$. For a comparison in which position $f$ is shown as Version A and position $g$ is shown as Version B, the first-shown candidate receives a presentation-order adjustment $e^\delta$. The model is
\begin{align}
\Pr(f \succ g) &=
\frac{v_f e^\delta}
{v_f e^\delta + v_g + \nu \sqrt{v_f v_g e^\delta}}, \nonumber\\
\Pr(g \succ f) &=
\frac{v_g}
{v_f e^\delta + v_g + \nu \sqrt{v_f v_g e^\delta}}, \nonumber\\
\Pr(\mathrm{tie}) &=
\frac{\nu \sqrt{v_f v_g e^\delta}}
{v_f e^\delta + v_g + \nu \sqrt{v_f v_g e^\delta}},
\label{eq:app-davidson}
\end{align}
where $\nu\ge 0$ controls the propensity to declare a tie and $\delta$ captures any systematic presentation-order effect. We report normalized position values
\[
v_k^\ast = \frac{v_k}{\sum_{j=1}^4 v_j},
\]
so that the four values form a preference-share vector over the four positions.

The split-half analysis asks whether these aggregate position preferences are stable under resampling. We randomly partition the human-evaluated source instances into two disjoint halves, re-estimate the model in \Cref{eq:app-davidson} on each half, and obtain two normalized preference vectors,
\[
\mathbf{v}^{\ast,(1)}=(v_1^{\ast,(1)},\ldots,v_4^{\ast,(1)})
\quad\text{and}\quad
\mathbf{v}^{\ast,(2)}=(v_1^{\ast,(2)},\ldots,v_4^{\ast,(2)}).
\]
For each split, we compute the correlation between the two vectors. Because each vector is estimated from only half of the data, we apply the Spearman-Brown correction \citep{spearman1910correlation},
\[
\rho_{\mathrm{SB}}=\frac{2\rho}{1+\rho},
\]
where $\rho$ is the split-half correlation. 
We repeat this procedure for $B=100$ random splits and report the corrected reliability.

This analysis measures the stability of the aggregate placement signal used for benchmark evaluation. The unit of reliability is the estimated preference vector over positions, matching the target used to evaluate model judges. The unlabeled condition shows high reliability across all six rubric dimensions, with corrected split-half reliability ranging from $0.952$ to $0.999$. Overall preference is also stable in the labeled condition, with reliability $0.975$. The reliabilities are 0.983, 0.998, 0.991, 0.992 and 0.985 for the first five dimensions, respectively.

\begin{table}[h]
\centering
\setlength{\tabcolsep}{6pt}
\renewcommand{\arraystretch}{1.05}
\begin{tabular}{llc}
\toprule
Condition & Dimension & Split-half reliability \\
\midrule
Labeled & Naturalness & 0.983 \\
Labeled & Visibility & 0.998 \\
Labeled & Helpfulness & 0.991 \\
Labeled & Appropriateness & 0.992 \\
Labeled & Click likelihood & 0.985 \\
Labeled & Overall preference & 0.975 \\
\midrule
Unlabeled & Naturalness & 0.980 \\
Unlabeled & Visibility & 0.999 \\
Unlabeled & Helpfulness & 0.952 \\
Unlabeled & Appropriateness & 0.983 \\
Unlabeled & Click likelihood & 0.957 \\
Unlabeled & Overall preference & 0.966 \\
\bottomrule
\end{tabular}
\caption{Spearman-Brown-corrected split-half reliability of aggregate position preferences. Each estimate is computed from $B=100$ random splits of the human-evaluated source instances.}
\label{tab:split-half-reliability}
\end{table}



\section{Omitted details in \Cref{sec:evaluation-protocol}}\label{app:evaluation-details}

This section further specifies the evaluation mechanics like A/B/T prediction, the model-judge prompt, placement-score estimation, presentation-swap error, and agreement among model judges.
\subsection{A/B/T Prediction}\label{app:prediction-details}

Each evaluation unit is a tuple $(x,c,p,q)$, where $x$ is a source instance, $c$ is the disclosure condition, $p$ is one of the six unordered placement pairs for that instance, and $q\in\{Q1,\ldots,Q6\}$ is one of the six benchmark target dimensions. The label space is
\[
\mathcal{Y}=\{\mathrm{A},\mathrm{B},\mathrm{Tie}\}.
\]
The reference label $y_{x,c,p,q}\in\mathcal{Y}$ is the retained human annotation after quality control. The auxiliary placement-attention item used in the human interface is not a benchmark target and is excluded from prediction accuracy.

For a test set $\mathcal{D}_{\mathrm{test}}$, exact A/B/T accuracy is
\[
\mathrm{Acc}
=
\frac{1}{|\mathcal{D}_{\mathrm{test}}|}
\sum_{(x,c,p,q)\in\mathcal{D}_{\mathrm{test}}}
\mathbf{1}\!\left[
\hat{y}_{x,c,p,q}=y_{x,c,p,q}
\right],
\]
where $\hat{y}_{x,c,p,q}$ is the model prediction. The metric gives credit only for exact agreement. Predicting $\mathrm{A}$ or $\mathrm{B}$ for a reference $\mathrm{Tie}$, or predicting $\mathrm{Tie}$ for a directional reference label, is counted as an error.

For the fine-tuned judges, train and test partitions are constructed at the source-instance level. All comparisons and target dimensions derived from the same source instance are assigned to the same partition to avoid information leakage. 
\subsection{Model-Judge Prompt}\label{app:model-judge-prompt}

Model judges are prompted to answer all six benchmark target dimensions in one JSON object. 
{\bf System Prompt}
\begin{Verbatim}[breaklines=true, breaksymbol=]
You are a careful evaluator comparing two versions of the same assistant response.
The two versions may differ in where an advertisement of the same product appears.
Your task is to evaluate how the inserted advertisement affects the response in different dimensions.

You must answer exactly six comparison questions.
You must choose A, B, or Tie for each question.

Do not write step-by-step analysis.
Return only the final JSON object.
Keep each reasoning field under 15 words.
Do not output markdown or any text outside the JSON.
\end{Verbatim}

{\bf User Prompt}
\begin{Verbatim}[breaklines=true, breaksymbol=]
You will compare two versions of the same assistant response.

Both versions contain an inserted advertisement of the same product.
Please pay close attention to the advertisement sentence and evaluate how it affects the response.

You are given:
1. The previous conversation context, if available.
2. The user's current question.
3. Version A.
4. Version B.

Your task is to answer six comparison questions.

Questions:
Q1. Naturalness of Flow - Choose the version that reads more smoothly and feels less disruptive. A natural version fits the surrounding response and does not feel abruptly inserted.
Q2. Ad Visibility - Choose the version where the advertisement is more visible. This does not mean the version is better or worse; it only asks which advertisement is easier to notice.
Q3. Content Helpfulness - Choose the version that better preserves the usefulness of the non-ad content in answering the user's question. Focus on whether the main response stays useful, not on whether the advertisement itself is useful.
Q4. Click Likelihood - If you had to click on one advertisement, choose the version that would make you more likely to click.
Q5. Appropriateness - Choose the version that feels more trustworthy and less manipulative. Consider whether the advertisement feels suitable, transparent, and not overly pushy.
Q6. Overall Preference - Choose the version you prefer, all things considered.

Output rules:
- Output valid JSON only.
- Do not include markdown.
- Do not include any text before or after the JSON.
- Each choice must be exactly "A", "B", or "Tie".
- Choose "Tie" if Version A and Version B are equally good, similarly flawed, or if the difference between them is minor, subjective, or not clear.
- When you are uncertain whether A or B is better, choose "Tie".
- Reasoning must be one short phrase under 15 words.
- Do not write step-by-step analysis.
- Return the final JSON object immediately.
- Use this exact schema:

{
  "Q1": {
    "choice": "A",
    "reasoning": "Brief reason."
  },
  "Q2": {
    "choice": "Tie",
    "reasoning": "Brief reason."
  },
  "Q3": {
    "choice": "B",
    "reasoning": "Brief reason."
  },
  "Q4": {
    "choice": "A",
    "reasoning": "Brief reason."
  },
  "Q5": {
    "choice": "A",
    "reasoning": "Brief reason."
  },
  "Q6": {
    "choice": "B",
    "reasoning": "Brief reason."
  }
}

Previous context:
{context}

User question:
{question}

Version A:
{version_a}

Version B:
{version_b}
\end{Verbatim}

A model response is accepted only if it parses as JSON and contains valid choices for Q1-Q6. Choice strings are normalized to the canonical labels $\mathrm{A}$, $\mathrm{B}$, and $\mathrm{Tie}$; free-form rationale text is not used to infer a missing choice.

\subsection{Presentation-Swap Error}\label{app:swap-error}

The presentation-swap test evaluates whether a model judge gives the same underlying verdict when Version A and Version B are exchanged. Let $\hat{y}_{u}^{\mathrm{orig}}$ be the prediction for evaluation unit $u$ under the original presentation order, and let $\hat{y}_{u}^{\mathrm{swap}}$ be the prediction after swapping the two candidate responses.

The required label map is
\[
m(\mathrm{A})=\mathrm{B},\qquad
m(\mathrm{B})=\mathrm{A},\qquad
m(\mathrm{Tie})=\mathrm{Tie}.
\]
A prediction is swap-consistent when
\[
\hat{y}_{u}^{\mathrm{swap}}=m(\hat{y}_{u}^{\mathrm{orig}}).
\]
The swap error is
\[
\mathrm{SwapErr}
=
\frac{1}{|\mathcal{D}_{\mathrm{swap}}|}
\sum_{u\in\mathcal{D}_{\mathrm{swap}}}
\mathbf{1}\!\left[
\hat{y}_{u}^{\mathrm{swap}}
\neq
m(\hat{y}_{u}^{\mathrm{orig}})
\right].
\]
This definition counts failures after applying the necessary A/B reversal. Tie predictions are consistent only when they remain Tie after the swap.

\subsection{Agreement Among Model Judges}\label{app:judge-agreement}

We use chance-corrected agreement to compare different model judges. For two judges $a$ and $b$, let $y_a(u)$ and $y_b(u)$ denote their A/B/T predictions on unit $u$. The observed agreement is
\[
p_o^{a,b}
=
\frac{1}{N}
\sum_{u=1}^{N}
\mathbf{1}\!\left[y_a(u)=y_b(u)\right].
\]
Let
\[
\hat{\pi}_a(y)
=
\frac{1}{N}
\sum_{u=1}^{N}
\mathbf{1}\!\left[y_a(u)=y\right]
\]
be judge $a$'s empirical marginal frequency for label $y\in\mathcal{Y}$. The agreement expected from the two judges' marginal label distributions is
\[
p_e^{a,b}
=
\sum_{y\in\mathcal{Y}}\hat{\pi}_a(y)\hat{\pi}_b(y).
\]
Cohen's $\kappa$ \citep{cohen1960coefficient} is
\[
\kappa^{a,b}
=
\frac{p_o^{a,b}-p_e^{a,b}}{1-p_e^{a,b}}.
\]
We compute this quantity for every pair of model judges and summarize the resulting distribution across judge pairs. 

For panel-level agreement, we use Fleiss' $\kappa$ \citep{fleiss1971measuring}. Let $R$ be the number of model judges, and let $n_{u,y}$ be the number of judges assigning unit $u$ to label $y$. The agreement for unit $u$ is
\[
P_u
=
\frac{1}{R(R-1)}
\left(\sum_{y\in\mathcal{Y}} n_{u,y}^{2}-R\right).
\]
With
\[
\bar{P}
=
\frac{1}{N}\sum_{u=1}^{N}P_u
\quad\text{and}\quad
\bar{p}_y
=
\frac{1}{NR}
\sum_{u=1}^{N}n_{u,y},
\]
the chance agreement is
\[
P_e=\sum_{y\in\mathcal{Y}}\bar{p}_y^2,
\]
and Fleiss' $\kappa$ is
\[
\kappa_F
=
\frac{\bar{P}-P_e}{1-P_e}.
\]
These statistics measure agreement among model judges. Human aggregate reliability is evaluated separately with the split-half test in \Cref{app:human-reliability}.

\subsection{Paired Statistical Comparisons}\label{app:paired-statistical-comparisons}

When comparing two systems on the same evaluation units, we use McNemar's test. Let $n_{01}$ be the number of units on which system $a$ is incorrect and system $b$ is correct, and let $n_{10}$ be the number of units on which system $a$ is correct and system $b$ is incorrect. The continuity-corrected statistic is
\[
\chi^2
=
\frac{(|n_{01}-n_{10}|-1)^2}{n_{01}+n_{10}}.
\]
When a family of paired comparisons is reported, we adjust $p$-values using the Holm procedure. We use these paired tests to assess whether an observed accuracy advantage on \textsc{LLMAdBench} is statistically reliable after accounting for evaluation-sample variation and multiple comparisons.

\section{Omitted Details in \Cref{sec:benchmark-results}}\label{app:full-results}

We report the full experimental results here. All systems are evaluated on the same held-out test rows, with the same parsing rules and scoring rules. We only make superficial modifications to the Qwen3-8B prompt to conform to the training template. For fair comparison, we report the fixed epoch-5 DPO Qwen3-8B checkpoint in both disclosure conditions.

\subsection{Accuracy with Confidence Intervals}\label{app:accuracy-ci}

\Cref{tab:main-CI} reports held-out A/B/T prediction accuracy with 95\% bootstrap confidence intervals. Accuracy is computed over individual rubric questions, and each held-out placement pair contributes six A/B/T prediction targets, one for each dimension in the rubric.
\begin{table}[h]
\centering
\setlength{\tabcolsep}{5.6pt}
\renewcommand{\arraystretch}{1.06}
\begin{tabular}{lcc}
\toprule
System & Labeled & Unlabeled \\
\midrule
\multicolumn{3}{l}{\emph{Task-adapted judge}}\\
Qwen3-8B, pre-DPO
& $40.25${\scriptsize$\pm1.59$}
& $36.52${\scriptsize$\pm1.53$} \\
Qwen3-8B, post-DPO epoch 5
& $\mathbf{50.54}${\scriptsize$\pm{1.62}$}
& $\mathbf{55.76}${\scriptsize$\pm{1.59}$} \\
\midrule
\multicolumn{3}{l}{\emph{Zero-shot frontier judges}}\\
Gemini-3.1-Pro
& $48.14${\scriptsize$\pm1.60$}
& $49.92${\scriptsize$\pm1.60$} \\
GPT-5.4
& $47.85${\scriptsize$\pm1.60$}
& $50.32${\scriptsize$\pm1.61$} \\
Kimi-K3
& $47.23${\scriptsize$\pm1.60$}
& $49.87${\scriptsize$\pm1.61$} \\
Claude-Opus-4.6
& $46.96${\scriptsize$\pm1.60$}
& $48.74${\scriptsize$\pm1.60$} \\
Mistral-Large-3
& $46.39${\scriptsize$\pm1.60$}
& $48.76${\scriptsize$\pm1.61$} \\
Grok-4.1
& $46.31${\scriptsize$\pm1.62$}
& $47.52${\scriptsize$\pm1.61$} \\
Llama-4-Maverick
& $44.51${\scriptsize$\pm1.59$}
& $48.09${\scriptsize$\pm1.59$} \\
DeepSeek-V4-Pro
& $44.00${\scriptsize$\pm1.60$}
& $46.18${\scriptsize$\pm1.59$} \\
\bottomrule
\end{tabular}
\caption{Held-out A/B/T prediction accuracy with 95\% bootstrap confidence intervals. The table uses the final eight-model benchmark comparison and the fixed epoch-5 DPO checkpoint. Intervals are reported as half-widths.}
\label{tab:main-CI}
\end{table}



\subsection{Paired Statistical Comparisons}\label{app:benchmark-paired-tests}

We report paired McNemar tests comparing the fixed epoch-5 DPO checkpoint against each zero-shot frontier judge in \Cref{tab:mcnemar}. Holm correction is applied separately within each disclosure condition over the eight frontier-judge comparisons. 


\begin{table}[!ht]
\centering
\begin{tabular}{llrrrr}
\toprule
Regime & Comparator & $\Delta$ (pp) & $\chi^2$ & $p$ & $p_{\text{adj}}$ \\
\midrule
\multirow{8}{*}{Labeled}
 & Gemini-3.1-Pro        & $-2.40$ & $6.9$   & $0.0088$ & $0.0088$ \\
 & GPT-5.4               & $-2.69$ & $8.2$   & $0.0041$ & $0.0083$ \\
 & Kimi-K3               & $-3.31$ & $11.9$  & $5.6\times10^{-4}$ & $1.7\times10^{-3}$ \\
 & Claude-Opus-4.6       & $-3.58$ & $14.1$  & $1.7\times10^{-4}$ & $6.8\times10^{-4}$ \\
 & Mistral-Large-3       & $-4.15$ & $19.8$  & $8.6\times10^{-6}$ & $5.2\times10^{-5}$ \\
 & Grok-4.1              & $-4.23$ & $19.4$  & $1.1\times10^{-5}$ & $5.3\times10^{-5}$ \\
 & Llama-4-Maverick      & $-6.03$ & $35.6$  & $2.4\times10^{-9}$ & $1.7\times10^{-8}$ \\
 & DeepSeek-V4-Pro       & $-6.54$ & $46.5$  & $9.4\times10^{-12}$ & $7.5\times10^{-11}$ \\
\midrule
\multirow{8}{*}{Unlabeled}
 & GPT-5.4               & $-5.44$  & $26.4$  & $2.7\times10^{-7}$ & $2.7\times10^{-7}$ \\
 & Gemini-3.1-Pro        & $-5.84$  & $36.5$  & $1.6\times10^{-9}$ & $4.6\times10^{-9}$ \\
 & Kimi-K3               & $-5.89$  & $33.6$  & $6.7\times10^{-9}$ & $1.4\times10^{-8}$ \\
 & Mistral-Large-3       & $-7.00$  & $49.4$  & $2.1\times10^{-12}$ & $8.3\times10^{-12}$ \\
 & Claude-Opus-4.6       & $-7.02$  & $50.3$  & $1.3\times10^{-12}$ & $6.4\times10^{-12}$ \\
 & Llama-4-Maverick      & $-7.67$  & $57.6$  & $3.3\times10^{-14}$ & $2.0\times10^{-13}$ \\
 & Grok-4.1              & $-8.23$  & $64.5$  & $9.7\times10^{-16}$ & $6.8\times10^{-15}$ \\
 & DeepSeek-V4-Pro       & $-9.58$  & $83.3$  & $7.2\times10^{-20}$ & $5.7\times10^{-19}$ \\
\bottomrule
\end{tabular}
\caption{McNemar tests against \textsc{Post-DPO} on paired predictions, full
evaluation sets. $\Delta$ is the comparator's accuracy minus ours, in points;
negative means our model is better. $p_{\text{adj}}$ is Holm-corrected within
each regime over the eight comparisons. \textsc{Post-DPO} is the epoch-$5$
checkpoint in both regimes. The adapted model beats all eight baselines in both
regimes; on unlabeled, every margin is large while on labeled the two smallest are
modest.}
\label{tab:mcnemar}
\end{table}

\subsection{DPO Training Setup and Checkpoint Accuracy}
\label{app:dpo-training-setup}

We train Qwen3-8B with Direct Preference Optimization using \texttt{trl}~\citep{vonwerra2020trl}. 
The preference data are derived from the same A/B/T benchmark labels used for evaluation. For comparisons with a clear preference, the selected version is treated as the chosen response, while both the non-selected version and the tie response are included as rejected responses. When annotators select a tie, the tie response is treated as chosen, and both Version A and Version B are treated as rejected. We further augment the data by swapping the presentation order of Versions A and B and reversing the corresponding A/B label. To avoid overrepresenting tie-versus-version comparisons introduced by this expansion, we randomly subsample these comparisons to maintain a balanced preference dataset. We construct the train/test split at the original question level, before any of these preference-data augmentations, using an 80/20 split. This prevents augmented variants of the same underlying comparison, for example, the original and A/B-swapped versions, from appearing in both the training and test sets. For the labeled condition, the resulting preference dataset contains 23,530 training rows and 3,714 test rows. For the unlabeled condition, it contains 21,381 training rows and 3,716 test rows.


We report the main training hyperparameters in \Cref{tab:dpo-training-hparams}. We use LoRA adaptation rather than full-parameter fine-tuning. The DPO loss uses the sigmoid objective with $\beta=0.1$. Optimization uses AdamW with a learning rate of $5\times10^{-6}$, cosine decay, and a warmup ratio of $0.03$. We train for five epochs on one NVIDIA H200 GPU.

\begin{table}[h]
\centering
\setlength{\tabcolsep}{6pt}
\renewcommand{\arraystretch}{1.06}
\begin{tabular}{ll}
\toprule
Component & Setting \\
\midrule
LoRA rank  & $32$ \\
LoRA alpha  &  $64$ \\
LoRA  dropout &  $0.05$ \\
Maximum sequence length & $4096$ \\
Batch size & $4$ \\
Gradient accumulation steps & $8$ \\
Training epochs & $5$ \\
Seed & $42$ \\
\bottomrule
\end{tabular}
\caption{Main DPO training hyperparameters for Qwen3-8B.}
\label{tab:dpo-training-hparams}
\end{table}

\begin{figure}[!htp]
\centering
\includegraphics[width=0.92\linewidth]{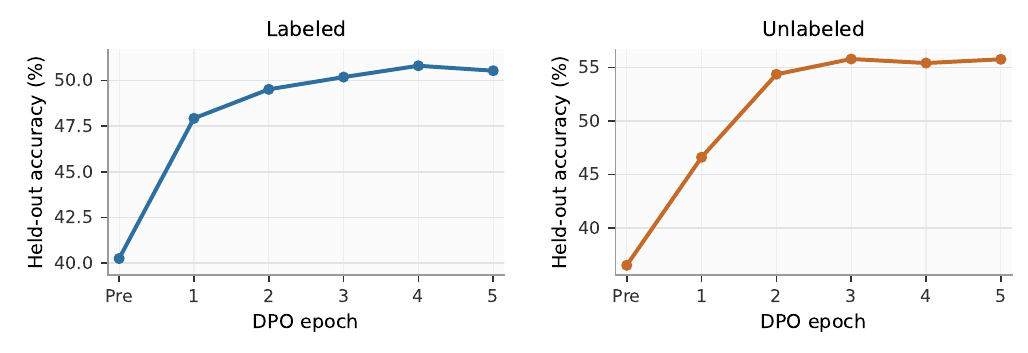}
\caption{Held-out A/B/T prediction accuracy across DPO checkpoints for labeled and unlabeled conditions. Main results use the epoch-5 checkpoint in both conditions; intermediate checkpoints are shown only for auditability.}
\label{fig:dpo-training-curves}
\end{figure}

\Cref{fig:dpo-training-curves} shows held-out A/B/T prediction accuracy across DPO checkpoints. Most of the gain appears in the first two epochs. In the labeled condition, accuracy increases from $40.25$ before DPO to $47.93$ after one epoch, $49.52$ after two epochs, $50.19$ after three epochs, $50.81$ after four epochs, and $50.54$ after five epochs. In the unlabeled condition, accuracy increases from $36.52$ before DPO to $46.61$ after one epoch, $54.36$ after two epochs, $55.79$ after three epochs, $55.41$ after four epochs, and $55.76$ after five epochs. The curve is therefore largely saturated by the second or third epoch, with only small changes afterward. The main leaderboard reports the epoch-5 checkpoint for both conditions.

\subsection{Error Breakdown by Label and Rubric Dimension}
\label{app:dpo-error-breakdown}

The main leaderboard averages over all A/B/T prediction targets. We further decompose performance by gold label and by rubric dimension in \Cref{tab:dpo-recall-by-label-app,tab:dpo-accuracy-by-dimension-app}. 


\begin{table*}[!ht]
\centering
\setlength{\tabcolsep}{4pt}
\renewcommand{\arraystretch}{1.04}
\begin{tabular}{lccc ccc}
\toprule
& \multicolumn{3}{c}{Labeled} & \multicolumn{3}{c}{Unlabeled} \\
\cmidrule(lr){2-4}\cmidrule(lr){5-7}
Gold label & A & B & Tie & A & B & Tie \\
\midrule
Qwen3-8B, pre-DPO & 20.22 & 77.89 & 5.82 &12.21 & 66.69 & 22.33 \\
Qwen3-8B, post-DPO epoch 5 & 49.36 & 67.07 &20.50 & 49.55 &63.31 &52.52 \\
\midrule
GPT-5.4 & 55.04 & 63.29 & 3.57 & 55.00 & 65.06 & 3.30 \\
Claude-Opus-4.6 & 45.91 & 61.73 & 20.11 & 46.10 & 58.44 & 31.60 \\
Gemini-3.1-Pro & 44.42 & 62.95 & 26.46 & 40.26 & 64.42 & 38.21 \\
Grok-4.1 & 49.83 & 55.04 & 22.35 & 44.42 & 55.71 & 35.22 \\
Kimi-K3 & 51.12 & 60.72 & 13.23 & 49.48 & 59.09 & 28.46 \\
DeepSeek-V4-Pro & 45.91 & 60.78 & 7.41 & 47.79 & 55.58 & 19.50 \\
Llama-4-Maverick & 52.20 & 45.03 & 28.44 & 56.49 & 48.25 & 27.36 \\
Mistral-Large-3 & 48.88 & 63.69 & 7.67 & 46.82 & 62.86 & 19.34 \\
\bottomrule
\end{tabular}
\caption{Recall by gold A/B/T label on the prediction task.}
\label{tab:dpo-recall-by-label-app}
\end{table*}

The label-wise breakdown shows where the DPO gain comes from. Before preference training, Qwen3-8B exhibits strong asymmetry in that it recalls Version B much more often than Version A and rarely recovers Tie, especially in the labeled condition. DPO makes the prediction behavior substantially more balanced. In the labeled condition, Version-A recall rises from $20.22$ to $49.36$, and Tie recall rises from $5.82$ to $20.50$, while Version-B recall remains high at $67.07$. In the unlabeled condition, the same pattern is stronger, where Version-A recall rises from $12.21$ to $49.55$, and Tie recall rises from $22.33$ to $52.52$. Thus, the improvement in the main leaderboard is not just a shift toward a one-directional answer. It comes from recovering preference cases that the base model largely misses.

\begin{table}[!ht]
\centering
\footnotesize
\setlength{\tabcolsep}{5pt}
\renewcommand{\arraystretch}{1.04}
\begin{tabular}{lcccccc}
\toprule
System & Q1 & Q2 & Q3 & Q4 & Q5 & Q6 \\
\midrule
\multicolumn{7}{l}{\emph{Labeled}} \\
Qwen3-8B, pre-DPO  &44.24 &44.52 &31.71 &43.00 &39.63 &38.73 \\
Qwen3-8B, post-DPO epoch 5 &53.58 &53.87 &40.40 &50.17 &52.32 &53.64 \\
\midrule
GPT-5.4 & 56.70 & 49.19 & 39.33 & 37.50 & 49.85 & 55.09 \\
Claude-Opus-4.6 & 55.30 & 51.77 & 41.62 & 34.17 & 47.68 & 51.27 \\
Gemini-3.1-Pro & 55.61 & 49.35 & 38.41 & 44.83 & 48.61 & 52.73 \\
Grok-4.1 & 55.14 & 48.55 & 41.01 & 31.33 & 49.38 & 52.55 \\
Kimi-K3 & 54.98 & 48.06 & 41.62 & 38.83 & 47.21 & 53.09 \\
DeepSeek-V4-Pro & 48.44 & 46.94 & 35.98 & 35.50 & 47.52 & 50.18 \\
Llama-4-Maverick & 50.31 & 49.03 & 35.82 & 36.00 & 45.82 & 50.73 \\
Mistral-Large-3 & 53.43 & 50.16 & 33.54 & 41.17 & 47.37 & 53.82 \\
\midrule
\multicolumn{7}{l}{\emph{Unlabeled}} \\
Qwen3-8B, pre-DPO &41.48 &32.05 &32.75 &42.41 &35.98 &33.90 \\
Qwen3-8B, post-DPO epoch 5 &61.41 &70.30 &41.85 &45.98 &57.17&59.08\\
\midrule
GPT-5.4 & 52.25 & 55.70 & 38.50 & 47.68 & 54.52 & 53.77 \\
Claude-Opus-4.6 & 54.50 & 59.90 & 39.30 & 36.07 & 50.62 & 53.25 \\
Gemini-3.1-Pro & 51.29 & 64.77 & 36.10 & 45.20 & 52.80 & 50.17 \\
Grok-4.1 & 48.07 & 58.22 & 35.78 & 41.02 & 50.78 & 52.23 \\
Kimi-K3 & 52.73 & 58.39 & 37.22 & 44.12 & 52.34 & 55.31 \\
DeepSeek-V4-Pro & 45.82 & 51.68 & 37.54 & 42.26 & 49.69 & 50.68 \\
Llama-4-Maverick & 52.09 & 58.39 & 31.95 & 43.19 & 50.31 & 53.60 \\
Mistral-Large-3 & 53.54 & 53.86 & 36.58 & 44.12 & 51.40 & 53.77 \\
\bottomrule
\end{tabular}
\caption{Held-out A/B/T accuracy by rubric dimension.}
\label{tab:dpo-accuracy-by-dimension-app}
\end{table}
 
The dimension-wise breakdown gives the same picture from the rubric side. Post-DPO Qwen3-8B improves over its pre-DPO version on every rubric dimension in both disclosure conditions. The gain is especially large in the unlabeled condition, where accuracy increases by more than twenty points on Q2, Q5, and Q6. The labeled condition is more mixed against the frontier judges, namely, post-DPO Qwen3-8B is strongest on Q2, Q4, and Q5, while some zero-shot judges remain competitive on Q1 and Q6.

\section{Omitted Details in \Cref{sec:judge-reliability}}
\label{app:judge-reliability-full}
\subsection{Presentation Order and Swap Diagnostics}
\label{app:presentation-order-full}

We show presentation-swap error by judge, condition, and rubric dimension in \Cref{tab:swap-error-full}. Swap error is the percentage of predictions that do not preserve the required A/B/T verdict after Version A and Version B are exchanged.

\begin{table*}[!ht]
\centering
\setlength{\tabcolsep}{4pt}
\renewcommand{\arraystretch}{1.04}
\begin{tabular}{llccccccc}
\toprule
Condition & Judge & Nat. & Vis. & Help. & Click & Appr. & Overall & Mean \\
\midrule
Labeled & GPT-5.4 & 12.67 & 17.65 & 13.96 & 51.58 & 18.03 & 13.12 & 21.17 \\
Labeled & Grok-4.1 & 15.58 & 12.02 & 25.27 & 43.63 & 23.53 & 18.29 & 23.06 \\
Labeled & Kimi-K3 & 15.77 & 22.82 & 28.44 & 46.28 & 14.16 & 14.61 & 23.68 \\
Labeled & Gemini-3.1-Pro & 31.42 & 13.90 & 25.86 & 38.46 & 27.08 & 28.05 & 27.46 \\
Labeled & Claude-Opus-4.6 & 20.30 & 25.40 & 40.27 & 43.57 & 25.34 & 24.05 & 29.82 \\
Labeled & Mistral-Large-3 & 34.58 & 23.27 & 36.78 & 30.90 & 32.19 & 32.97 & 31.78 \\
Labeled & Llama-4-Maverick & 36.26 & 31.16 & 33.35 & 31.67 & 35.62 & 31.93 & 33.33 \\
Labeled & DeepSeek-V4-Pro & 44.54 & 29.73 & 47.45 & 47.25 & 38.20 & 39.56 & 41.12 \\
\midrule
Labeled & Mean & 26.39 & 21.99 & 31.42 & 41.67 & 26.77 & 25.32 & 28.93 \\
\midrule
Unlabeled & Grok-4.1 & 21.64 & 16.54 & 14.08 & 45.35 & 30.17 & 26.16 & 25.66 \\
Unlabeled & GPT-5.4 & 23.26 & 22.16 & 25.97 & 48.64 & 23.39 & 23.26 & 27.78 \\
Unlabeled & Gemini-3.1-Pro & 35.14 & 19.32 & 12.27 & 38.44 & 32.24 & 32.75 & 28.36 \\
Unlabeled & Kimi-K3 & 22.80 & 25.00 & 30.17 & 48.32 & 21.77 & 22.87 & 28.49 \\
Unlabeled & Claude-Opus-4.6 & 26.61 & 24.29 & 40.57 & 46.71 & 27.00 & 27.39 & 32.10 \\
Unlabeled & Llama-4-Maverick & 49.48 & 37.40 & 32.30 & 35.27 & 42.38 & 43.48 & 40.05 \\
Unlabeled & Mistral-Large-3 & 42.83 & 37.47 & 44.64 & 39.86 & 42.18 & 42.38 & 41.56 \\
Unlabeled & DeepSeek-V4-Pro & 50.65 & 37.79 & 56.07 & 47.87 & 44.77 & 44.77 & 46.99 \\
\midrule
Unlabeled & Mean & 34.05 & 27.50 & 32.01 & 43.81 & 32.99 & 32.88 & 33.87 \\
\bottomrule
\end{tabular}
\caption{Presentation-swap error (\%) for the eight zero-shot LLM judges. Lower values indicate greater invariance to exchanging Version A and Version B.}
\label{tab:swap-error-full}
\end{table*}

\Cref{tab:judge-slot-full} on the other hand, reports the Version-A order term estimated for overall preference. This diagnostic complements the swap test. Swap error records whether a judge changes its underlying A/B/T verdict after the two versions are exchanged; the order term estimates whether the judge's fitted choice model assigns residual preference to Version A after accounting for the two candidate placement positions.

\begin{table}[!ht]
\centering
\setlength{\tabcolsep}{4pt}
\renewcommand{\arraystretch}{1.04}
\begin{tabular}{llccc}
\toprule
Judge & Condition & $P(\mathrm{A})$ & Version-A $\delta$ [95\% CI] & Sig. \\
\midrule
Grok-4.1 & Labeled & 0.473 & $-0.19$ $[-0.25, -0.13]$ & 12/12 \\
 & Unlabeled & 0.437 & $-0.28$ $[-0.34, -0.22]$  &  \\
\addlinespace[2pt]
GPT-5.4 &Labeled & 0.494 & $-0.04$ $[-0.11, +0.01]$ & 7/12 \\
 &Unlabeled & 0.490 & $-0.05$ $[-0.12, +0.04]$  &  \\
\addlinespace[2pt]
Kimi-K3 & Labeled & 0.472 & $-0.19$ $[-0.25, -0.12]$& 10/12 \\
 & Unlabeled & 0.483 & $-0.10$ $[-0.18, -0.02]$  &  \\
\addlinespace[2pt]
Gemini-3.1-Pro & Labeled & 0.372 & $-0.87$ $[-0.95, -0.79]$ & 11/12 \\
 & Unlabeled & 0.353 & $-0.76$ $[-0.85, -0.69]$  &  \\
\addlinespace[2pt]
Claude-Opus-4.6 & Labeled & 0.416 & $-0.59$ $[-0.68, -0.52]$ & 11/12 \\
 & Unlabeled & 0.410 & $-0.67$ $[-0.77, -0.59]$  &  \\
\addlinespace[2pt]
Llama-4-Maverick & Labeled & 0.541 & $+0.24$ $[+0.13, +0.34]$ & 9/12 \\
 &Unlabeled & 0.589 & $+0.47$ $[+0.34, +0.59]$   &  \\
\addlinespace[2pt]
Mistral-Large-3 & Labeled & 0.386 & $-0.58$ $[-0.67, -0.50]$ & 11/12 \\
 & Unlabeled & 0.390 & $-0.53$ $[-0.64, -0.42]$  &  \\
\addlinespace[2pt]
DeepSeek-V4-Pro &Labeled & 0.357 & $-0.80$ $[-0.91, -0.73]$ & 12/12 \\
 & Unlabeled & 0.369 & $-0.65$ $[-0.72, -0.56]$   &  \\
\bottomrule
\end{tabular}
\caption{Version-A order term for overall preference. $P(\mathrm{A})$ is the probability that the judge chooses Version A among non-tie outcomes. The coefficient $\delta$ is estimated in the Davidson-Bradley-Terry model after accounting for the two candidate placement positions. Negative values indicate residual preference for Version B, and positive values indicate residual preference for Version A. The ``Sig.'' column counts dimension-by-condition cells, out of 12, whose pointwise 95\% cluster-bootstrap confidence interval excludes zero. }
\label{tab:judge-slot-full}
\end{table}

The estimated order term is not concentrated in most models. In the overall-preference rows, GPT-5.4 is the only judge whose confidence intervals include zero in both conditions. Most other judges have negative estimates, indicating residual preference for Version B after the fitted model accounts for the two placement positions. Llama-4-Maverick is the exception, with positive estimates in both conditions. This pattern supports our result that A/B presentation is not a harmless implementation detail for zero-shot LLM judges.

\subsection{Agreement Among LLM Judges}
\label{app:judge-agreement-full}

In \Cref{tab:judge-agreement-by-dimension}, we summarize agreement among the eight zero-shot judges on the original A/B presentation. Raw agreement here means model-model agreement on the same A/B/T item, not agreement with human annotators. Cohen's $\kappa$ is computed for every pair of judges and summarized over the 28 judge pairs, while Fleiss' $\kappa$ is computed over the full judge panel.

\begin{table*}[!ht]
\centering
\setlength{\tabcolsep}{4.5pt}
\renewcommand{\arraystretch}{1.04}
\begin{tabular}{llcccccc}
\toprule
\multirow{2}{*}{Condition}
& \multirow{2}{*}{Dimension}
& \multicolumn{4}{c}{Pairwise Cohen's $\kappa$}
& \multirow{2}{*}{Fleiss' $\kappa$}
& \multirow{2}{*}{Raw model-model agree.} \\
\cmidrule(lr){3-6}
& & Mean & Median & Min & Max & & \\
\midrule
Labeled & Naturalness & 0.452 & 0.410 & 0.232 & 0.702 & 0.451 & 0.722 \\
Labeled & Visibility & 0.574 & 0.548 & 0.460 & 0.727 & 0.572 & 0.765 \\
Labeled & Helpfulness & 0.280 & 0.274 & 0.087 & 0.543 & 0.256 & 0.507 \\
Labeled & Click likelihood & 0.143 & 0.139 & -0.132 & 0.407 & 0.136 & 0.465 \\
Labeled & Appropriateness & 0.447 & 0.459 & 0.212 & 0.643 & 0.445 & 0.708 \\
Labeled & Overall preference & 0.473 & 0.460 & 0.242 & 0.696 & 0.472 & 0.733 \\
\midrule
Labeled & All & 0.402 & 0.389 & 0.267 & 0.576 & 0.400 & 0.650 \\
\midrule
Unlabeled & Naturalness & 0.364 & 0.322 & 0.210 & 0.602 & 0.360 & 0.671 \\
Unlabeled & Visibility & 0.492 & 0.467 & 0.372 & 0.617 & 0.491 & 0.725 \\
Unlabeled & Helpfulness & 0.163 & 0.163 & 0.039 & 0.387 & 0.131 & 0.503 \\
Unlabeled & Click likelihood & 0.137 & 0.187 & -0.202 & 0.380 & 0.133 & 0.505 \\
Unlabeled & Appropriateness & 0.396 & 0.379 & 0.254 & 0.610 & 0.394 & 0.677 \\
Unlabeled & Overall preference & 0.395 & 0.363 & 0.228 & 0.628 & 0.392 & 0.689 \\
\midrule
Unlabeled & All & 0.384 & 0.360 & 0.278 & 0.546 & 0.384 & 0.628 \\
\bottomrule
\end{tabular}
\caption{Agreement among the eight zero-shot LLM judges on the original A/B presentation.}
\label{tab:judge-agreement-by-dimension}
\end{table*}

\Cref{tab:pairwise-kappa-matrix} gives the pairwise Cohen's $\kappa$ values behind the heatmap in \Cref{fig:judge-reliability-summary}. The pattern is structured rather than uniform. Grok-4.1, GPT-5.4, Kimi-K3, Gemini-3.1-Pro, and Claude-Opus-4.6 agree more strongly with one another than with Mistral-Large-3, Llama-4-Maverick, and DeepSeek-V4-Pro. It tells us that adding more LLM judges does not necessarily add independent measurements of the same preference target.

\begin{table*}[!ht]
\centering
\setlength{\tabcolsep}{4pt}
\renewcommand{\arraystretch}{1.04}
\begin{tabular}{lcccccccc}
\toprule
 & Grok & GPT-5.4 & Kimi & Gemini & Claude & Mistral & Llama-4 & DeepSeek \\
\midrule
Grok & - & 0.351 & 0.511 & 0.546 & 0.474 & 0.382 & 0.400 & 0.306 \\
GPT-5.4 & 0.453 & - & 0.465 & 0.300 & 0.422 & 0.334 & 0.278 & 0.320 \\
Kimi & 0.534 & 0.576 & - & 0.496 & 0.513 & 0.397 & 0.406 & 0.340 \\
Gemini & 0.526 & 0.398 & 0.468 & - & 0.496 & 0.356 & 0.363 & 0.280 \\
Claude & 0.520 & 0.502 & 0.528 & 0.486 & - & 0.358 & 0.307 & 0.320 \\
Mistral & 0.292 & 0.410 & 0.381 & 0.289 & 0.312 & - & 0.420 & 0.324 \\
Llama-4 & 0.320 & 0.336 & 0.339 & 0.345 & 0.267 & 0.375 & - & 0.297 \\
DeepSeek & 0.341 & 0.417 & 0.402 & 0.325 & 0.363 & 0.429 & 0.331 & - \\
\bottomrule
\end{tabular}
\caption{Pairwise Cohen's $\kappa$ among zero-shot LLM judges, matching \Cref{fig:judge-reliability-summary}. The lower triangle shows the labeled condition while the upper triangle shows the unlabeled condition. Values are computed on the original A/B presentation and pooled over Q1-Q6.}
\label{tab:pairwise-kappa-matrix}
\end{table*}

We also evaluate whether a judge panel can recover human labels by simple aggregation. For each held-out A/B/T item, the eight zero-shot judge predictions are combined by majority voting, and ties among plurality labels are resolved as Tie. This rule uses the same prediction targets as the main leaderboard for fair comparison. The resulting ensemble remains below the post-DPO Qwen3-8B judge, showing that the gain from human-preference training is not recovered by simply pooling more zero-shot judges. Specifically, paired McNemar tests show that the post-DPO Qwen3-8B judge outperforms majority voting in both conditions, with $\chi^2=8.4$ and $p=0.0037$ for labeled, and $\chi^2=29.8$ and $p=4.7\times 10^{-8}$ for unlabeled.

\begin{table}[h]
\centering
\small
\setlength{\tabcolsep}{6pt}
\renewcommand{\arraystretch}{1.05}
\begin{tabular}{lcc}
\toprule
System & Labeled & Unlabeled \\
\midrule
Best single zero-shot judge & {48.14 (Gemini-3.1-Pro)} &{50.32 (GPT-5.4)} \\
Eight-judge majority vote &{47.87} &{50.43} \\
Qwen3-8B post-DPO judge &{\bf 50.54} & {\bf 55.76} \\
\bottomrule
\end{tabular}
\caption{Held-out A/B/T accuracy of simple majority voting among the eight zero-shot LLM judges.}
\label{tab:judge-majority-vote}
\end{table}

Furthermore, we compute tie rates in  \Cref{tab:judge-tie-rates}. We use tie rates to audit how each judge uses the A/B/T label space. Large differences in tie usage change both raw agreement and chance-corrected agreement, so they provide useful context for interpreting the agreement results above.

\begin{table*}[!ht]
\centering
\setlength{\tabcolsep}{4pt}
\renewcommand{\arraystretch}{1.04}
\begin{tabular}{llcccccc}
\toprule
Condition & Judge & Nat. & Vis. & Help. & Click & Appr. & Overall \\
\midrule
Labeled & Grok-4.1 & 0.008 & 0.013 & 0.495 & 0.359 & 0.061 & 0.021 \\
Labeled & GPT-5.4 & 0.003 & 0.021 & 0.008 & 0.143 & 0.008 & 0.004 \\
Labeled & Kimi-K3 & 0.006 & 0.065 & 0.238 & 0.131 & 0.022 & 0.006 \\
Labeled & Gemini-3.1-Pro & 0.036 & 0.036 & 0.699 & 0.112 & 0.036 & 0.027 \\
Labeled & Claude-Opus-4.6 & 0.022 & 0.113 & 0.214 & 0.356 & 0.065 & 0.034 \\
Labeled & Mistral-Large-3 & 0.000 & 0.054 & 0.182 & 0.019 & 0.006 & 0.000 \\
Labeled & Llama-4-Maverick & 0.001 & 0.104 & 0.723 & 0.165 & 0.034 & 0.002 \\
Labeled & DeepSeek-V4-Pro & 0.040 & 0.034 & 0.181 & 0.138 & 0.030 & 0.025 \\
\midrule
Unlabeled & Grok-4.1 & 0.004 & 0.016 & 0.846 & 0.175 & 0.089 & 0.018 \\
Unlabeled & GPT-5.4 & 0.002 & 0.011 & 0.030 & 0.087 & 0.006 & 0.006 \\
Unlabeled & Kimi-K3 & 0.009 & 0.079 & 0.625 & 0.079 & 0.031 & 0.013 \\
Unlabeled & Gemini-3.1-Pro & 0.024 & 0.041 & 0.911 & 0.066 & 0.030 & 0.018 \\
Unlabeled & Claude-Opus-4.6 & 0.032 & 0.094 & 0.561 & 0.273 & 0.072 & 0.035 \\
Unlabeled & Mistral-Large-3 & 0.003 & 0.045 & 0.520 & 0.019 & 0.015 & 0.006 \\
Unlabeled & Llama-4-Maverick & 0.010 & 0.036 & 0.791 & 0.049 & 0.038 & 0.021 \\
Unlabeled & DeepSeek-V4-Pro & 0.069 & 0.034 & 0.379 & 0.119 & 0.050 & 0.044 \\
\bottomrule
\end{tabular}
\caption{Tie rates for zero-shot LLM judges on the original A/B presentation. Values are the fraction of predictions returned as Tie for each rubric dimension.}
\label{tab:judge-tie-rates}
\end{table*}

\subsection{Human Aggregate Reliability}
\label{app:human-aggregate-reliability-full}

\Cref{tab:split-half-reliability} further reports the split-half reliabilities used in \Cref{subsec:human-signal-stability}. The split is performed over source instances. For each random split, the placement model is fitted separately on the two halves, producing two normalized four-position preference vectors. Reliability is the Spearman-Brown-corrected correlation~\citep{spearman1910correlation} between the two vectors, averaged over $B=100$ random splits.


We end this section with \Cref{tab:annotator-profiles-full} which summarizes annotators' use of the A/B/T response options. The table reports the share of directional judgments assigned to Version A and the overall Tie rate for each disclosure condition, with between-annotator variation shown in parentheses.

\begin{table}[h]
\centering
\setlength{\tabcolsep}{6pt}
\renewcommand{\arraystretch}{1.05}
\begin{tabular}{lcccc}
\toprule
Condition & \# Annotators & Median comparisons & $P(\mathrm{A})$ & Tie rate \\
\midrule
Labeled & 32 & 49 & 0.508 {\scriptsize(0.053)} & 0.211 {\scriptsize(0.156)} \\
Unlabeled & 30 & 49.5 & 0.509 {\scriptsize(0.064)} & 0.165 {\scriptsize(0.137)} \\
\bottomrule
\end{tabular}
\caption{Descriptive annotator profiles. Parentheses report between-annotator standard deviations. $P(\mathrm{A})$ is computed among non-tie judgments; Version A corresponds to the left response in the annotation interface.}
\label{tab:annotator-profiles-full}
\end{table}

\section{Omitted Details in \Cref{sec:benchmark-findings}}
\label{app:placement-findings}

\subsection{Human Placement Scores Across Dimensions}
\label{app:human-placement-results}

We report human placement scores for all six rubric dimensions in \Cref{tab:human-placement-all-dimensions} with $B=300$ resamples. Naturalness, helpfulness, and appropriateness have the same top-position ordering as overall preference in both disclosure conditions. Visibility behaves differently, as expected from its definition. Front placement is the most salient position in both conditions, although explicit labeling makes interior advertisements easier to notice. Click likelihood also differs from the experience-oriented dimensions. Its mass is concentrated on the two edge positions, especially the end position.

\begin{table*}[!ht]
\centering
\setlength{\tabcolsep}{1.5pt}
\renewcommand{\arraystretch}{1.06}
\begin{tabular}{llccccc}
\toprule
Condition & Dimension & Front & Early & Late & End & $\delta$ \\
\midrule
Labeled & Naturalness
& 0.159 {\scriptsize[0.141, 0.182]}
& 0.191 {\scriptsize[0.165, 0.214]}
& 0.244 {\scriptsize[0.216, 0.277]}
& \textbf{0.406} {\scriptsize[0.370, 0.445]}
& $+0.034$ \\
Labeled & Visibility
& \textbf{0.605} {\scriptsize[0.561, 0.651]}
& 0.200 {\scriptsize[0.169, 0.231]}
& 0.105 {\scriptsize[0.086, 0.123]}
& 0.091 {\scriptsize[0.074, 0.109]}
& $+0.086$ \\
Labeled & Helpfulness
& 0.227 {\scriptsize[0.194, 0.255]}
& 0.163 {\scriptsize[0.139, 0.188]}
& 0.170 {\scriptsize[0.144, 0.204]}
& \textbf{0.440} {\scriptsize[0.399, 0.488]}
& $-0.062$ \\
Labeled & Click likelihood
& 0.304 {\scriptsize[0.269, 0.339]}
& 0.174 {\scriptsize[0.151, 0.200]}
& 0.153 {\scriptsize[0.129, 0.177]}
& \textbf{0.369} {\scriptsize[0.330, 0.416]}
& $+0.064$ \\
Labeled & Appropriateness
& 0.162 {\scriptsize[0.141, 0.185]}
& 0.183 {\scriptsize[0.162, 0.205]}
& 0.223 {\scriptsize[0.197, 0.250]}
& \textbf{0.432} {\scriptsize[0.395, 0.470]}
& $+0.018$ \\
Labeled & Overall preference
& 0.176 {\scriptsize[0.155, 0.197]}
& 0.209 {\scriptsize[0.182, 0.234]}
& 0.252 {\scriptsize[0.223, 0.287]}
& \textbf{0.364} {\scriptsize[0.329, 0.399]}
& $+0.070$ \\
\midrule
Unlabeled & Naturalness
& 0.125 {\scriptsize[0.109, 0.140]}
& 0.232 {\scriptsize[0.202, 0.262]}
& \textbf{0.380} {\scriptsize[0.341, 0.426]}
& 0.264 {\scriptsize[0.233, 0.296]}
& $+0.061$ \\
Unlabeled & Visibility
& \textbf{0.802} {\scriptsize[0.757, 0.847]}
& 0.046 {\scriptsize[0.033, 0.061]}
& 0.017 {\scriptsize[0.011, 0.024]}
& 0.136 {\scriptsize[0.106, 0.166]}
& $-0.027$ \\
Unlabeled & Helpfulness
& 0.165 {\scriptsize[0.141, 0.185]}
& 0.230 {\scriptsize[0.203, 0.261]}
& \textbf{0.339} {\scriptsize[0.302, 0.382]}
& 0.267 {\scriptsize[0.233, 0.299]}
& $+0.015$ \\
Unlabeled & Click likelihood
& 0.288 {\scriptsize[0.257, 0.321]}
& 0.188 {\scriptsize[0.167, 0.212]}
& 0.199 {\scriptsize[0.171, 0.223]}
& \textbf{0.325} {\scriptsize[0.293, 0.358]}
& $+0.056$ \\
Unlabeled & Appropriateness
& 0.117 {\scriptsize[0.100, 0.132]}
& 0.230 {\scriptsize[0.199, 0.260]}
& \textbf{0.374} {\scriptsize[0.338, 0.418]}
& 0.279 {\scriptsize[0.250, 0.314]}
& $+0.049$ \\
Unlabeled & Overall preference
& 0.132 {\scriptsize[0.115, 0.152]}
& 0.234 {\scriptsize[0.205, 0.264]}
& \textbf{0.352} {\scriptsize[0.317, 0.389]}
& 0.281 {\scriptsize[0.252, 0.313]}
& $+0.056$ \\
\bottomrule
\end{tabular}
\caption{Human Davidson-Bradley-Terry placement scores across all rubric dimensions. The order term $\delta$ is the Version-A presentation-order coefficient.}
\label{tab:human-placement-all-dimensions}
\end{table*}

\Cref{tab:human-disclosure-delta} then reports the point-estimate change from the unlabeled condition to the labeled condition. For overall preference, late placement loses $0.100$ score mass and end placement gains $0.083$, producing the late-to-end reversal reported in \Cref{subsec:disclosure-placement}. For visibility, the front score falls from $0.802$ to $0.605$, while early and late positions gain score mass. This indicates that the label itself makes within-response advertisements easier to identify. For helpfulness and appropriateness, end placement gains more than $0.15$ score mass, consistent with the interpretation that a labeled advertisement is less disruptive when placed after the answer. For click likelihood, the end position also gains score mass, while the late position falls.


\begin{table}[!ht]
\centering
\setlength{\tabcolsep}{5pt}
\renewcommand{\arraystretch}{1.06}
\begin{tabular}{lrrrr}
\toprule
Dimension & $\Delta$ Front & $\Delta$ Early & $\Delta$ Late & $\Delta$ End \\
\midrule
Naturalness & $+0.034$ & $-0.041$ & $-0.136$ & $+0.142$ \\
Visibility & $-0.197$ & $+0.154$ & $+0.088$ & $-0.045$ \\
Helpfulness & $+0.062$ & $-0.067$ & $-0.169$ & $+0.173$ \\
Click likelihood & $+0.016$ & $-0.014$ & $-0.046$ & $+0.044$ \\
Appropriateness & $+0.045$ & $-0.047$ & $-0.151$ & $+0.153$ \\
Overall preference & $+0.044$ & $-0.025$ & $-0.100$ & $+0.083$ \\
\bottomrule
\end{tabular}
\caption{Point-estimate disclosure effects in human placement scores. Each entry is the labeled score minus the unlabeled score for the same rubric dimension and position.}
\label{tab:human-disclosure-delta}
\end{table}

\subsection{LLM-Judge Placement and Dimension Profiles}
\label{app:llm-placement-results}

\Cref{tab:llm-overall-placement} gives the full overall-preference placement profiles for the eight zero-shot LLM judges used in the main benchmark comparison with $B=300$ resamples. We call a position modal when it has the largest placement score within a judge-condition row. The human modal position changes from late to end after labeling. The LLM judge profiles are more concentrated. Except for Mistral-Large-3 in the unlabeled condition, every retained zero-shot judge makes end placement modal in both disclosure conditions. The concentration is especially strong after labeling, where Claude-Opus-4.6, GPT-5.4, Gemini-3.1-Pro, and Kimi-K3 all assign more than $0.75$ of the overall-preference score mass to end placement.

\begin{table*}[!ht]
\centering
\setlength{\tabcolsep}{1.5pt}
\renewcommand{\arraystretch}{1.05}
\begin{tabular}{llccccc}
\toprule
Condition & Rater & Front & Early & Late & End & $\delta$ \\
\midrule
Labeled & Human
& 0.176 {\scriptsize[0.155, 0.197]}
& 0.209 {\scriptsize[0.182, 0.234]}
& 0.252 {\scriptsize[0.223, 0.287]}
& \textbf{0.364} {\scriptsize[0.329, 0.399]}
& $+0.070$ \\
\cmidrule(lr){1-7}
Labeled & Claude-Opus-4.6
& 0.056 {\scriptsize[0.041, 0.072]}
& 0.049 {\scriptsize[0.035, 0.066]}
& 0.049 {\scriptsize[0.035, 0.064]}
& \textbf{0.846} {\scriptsize[0.804, 0.884]}
& $-0.622^{*}$ \\
Labeled & DeepSeek-V4-Pro
& 0.067 {\scriptsize[0.053, 0.084]}
& 0.143 {\scriptsize[0.119, 0.169]}
& 0.133 {\scriptsize[0.110, 0.160]}
& \textbf{0.657} {\scriptsize[0.611, 0.701]}
& $-0.774^{*}$ \\
Labeled & Gemini-3.1-Pro
& 0.075 {\scriptsize[0.057, 0.099]}
& 0.054 {\scriptsize[0.040, 0.071]}
& 0.049 {\scriptsize[0.036, 0.065]}
& \textbf{0.821} {\scriptsize[0.777, 0.862]}
& $-0.924^{*}$ \\
Labeled & GPT-5.4
& 0.044 {\scriptsize[0.030, 0.056]}
& 0.059 {\scriptsize[0.041, 0.074]}
& 0.059 {\scriptsize[0.042, 0.079]}
& \textbf{0.838} {\scriptsize[0.797, 0.881]}
& $-0.018$ \\
Labeled & Grok-4.1
& 0.127 {\scriptsize[0.102, 0.155]}
& 0.085 {\scriptsize[0.066, 0.103]}
& 0.068 {\scriptsize[0.053, 0.085]}
& \textbf{0.720} {\scriptsize[0.669, 0.771]}
& $-0.176^{*}$ \\
Labeled & Kimi-K3
& 0.086 {\scriptsize[0.066, 0.106]}
& 0.069 {\scriptsize[0.052, 0.088]}
& 0.077 {\scriptsize[0.058, 0.096]}
& \textbf{0.768} {\scriptsize[0.721, 0.821]}
& $-0.245^{*}$ \\
Labeled & Llama-4-Maverick
& 0.053 {\scriptsize[0.043, 0.064]}
& 0.198 {\scriptsize[0.169, 0.225]}
& 0.250 {\scriptsize[0.218, 0.285]}
& \textbf{0.499} {\scriptsize[0.457, 0.543]}
& $+0.216^{*}$ \\
Labeled & Mistral-Large-3
& 0.079 {\scriptsize[0.068, 0.091]}
& 0.211 {\scriptsize[0.178, 0.248]}
& 0.250 {\scriptsize[0.215, 0.285]}
& \textbf{0.460} {\scriptsize[0.410, 0.509]}
& $-0.540^{*}$ \\
\midrule
Unlabeled & Human
& 0.132 {\scriptsize[0.115, 0.152]}
& 0.234 {\scriptsize[0.205, 0.264]}
& \textbf{0.352} {\scriptsize[0.317, 0.389]}
& 0.281 {\scriptsize[0.252, 0.313]}
& $+0.056$ \\
\cmidrule(lr){1-7}
Unlabeled & Claude-Opus-4.6
& 0.035 {\scriptsize[0.026, 0.043]}
& 0.101 {\scriptsize[0.080, 0.123]}
& 0.176 {\scriptsize[0.140, 0.206]}
& \textbf{0.688} {\scriptsize[0.641, 0.740]}
& $-0.675^{*}$ \\
Unlabeled & DeepSeek-V4-Pro
& 0.095 {\scriptsize[0.081, 0.112]}
& 0.169 {\scriptsize[0.147, 0.201]}
& 0.248 {\scriptsize[0.209, 0.283]}
& \textbf{0.488} {\scriptsize[0.443, 0.538]}
& $-0.609^{*}$ \\
Unlabeled & Gemini-3.1-Pro
& 0.100 {\scriptsize[0.082, 0.118]}
& 0.137 {\scriptsize[0.110, 0.159]}
& 0.176 {\scriptsize[0.146, 0.207]}
& \textbf{0.587} {\scriptsize[0.540, 0.639]}
& $-0.840^{*}$ \\
Unlabeled & GPT-5.4
& 0.079 {\scriptsize[0.064, 0.090]}
& 0.136 {\scriptsize[0.113, 0.160]}
& 0.236 {\scriptsize[0.198, 0.275]}
& \textbf{0.549} {\scriptsize[0.505, 0.601]}
& $-0.099$ \\
Unlabeled & Grok-4.1
& 0.162 {\scriptsize[0.138, 0.187]}
& 0.174 {\scriptsize[0.149, 0.204]}
& 0.260 {\scriptsize[0.227, 0.297]}
& \textbf{0.405} {\scriptsize[0.365, 0.449]}
& $-0.335^{*}$ \\
Unlabeled & Kimi-K3
& 0.067 {\scriptsize[0.057, 0.077]}
& 0.154 {\scriptsize[0.131, 0.182]}
& 0.177 {\scriptsize[0.146, 0.208]}
& \textbf{0.602} {\scriptsize[0.554, 0.648]}
& $-0.135^{*}$ \\
Unlabeled & Llama-4-Maverick
& 0.070 {\scriptsize[0.055, 0.082]}
& 0.204 {\scriptsize[0.177, 0.235]}
& 0.351 {\scriptsize[0.312, 0.384]}
& \textbf{0.376} {\scriptsize[0.334, 0.414]}
& $+0.405^{*}$ \\
Unlabeled & Mistral-Large-3
& 0.084 {\scriptsize[0.067, 0.104]}
& 0.215 {\scriptsize[0.190, 0.243]}
& \textbf{0.371} {\scriptsize[0.330, 0.413]}
& 0.330 {\scriptsize[0.298, 0.368]}
& $-0.592^{*}$ \\
\bottomrule
\end{tabular}
\caption{Davidson-Bradley-Terry placement scores for overall preference. The order term $\delta$ is the Version-A presentation-order coefficient. A star indicates that the pointwise 95\% cluster-bootstrap confidence interval for $\delta$ excludes zero.}
\label{tab:llm-overall-placement}
\end{table*}

We may also ask whether the version preferred on each rubric dimension is also the version preferred overall. \Cref{tab:dimension-overall-association-app} reports, after excluding ties on either question, how often each rubric dimension selects the same version as overall preference. Experience-oriented dimensions align closely with overall preference. Visibility goes in the opposite direction, where the version whose advertisement is easier to notice is usually not the version preferred overall. Click likelihood lies between these cases for humans, while zero-shot model judges show much less consistent behavior.

\begin{table}[!ht]
\centering
\setlength{\tabcolsep}{4.5pt}
\renewcommand{\arraystretch}{1.04}
\begin{tabular}{llccccc}
\toprule
Condition & Rater & Nat. & Help. & Appr. & Click & Vis. \\
\midrule
Unlabeled & Human & 0.97 & 0.96 & 0.97 & 0.75 & 0.30 \\
Unlabeled & Claude-Opus-4.6 & 0.97 & 0.94 & 0.98 & 0.64 & 0.12 \\
Unlabeled & DeepSeek-V4-Pro & 0.95 & 0.95 & 0.97 & 0.28 & 0.11 \\
Unlabeled & Gemini-3.1-Pro & 0.99 & 1.00 & 0.99 & 0.96 & 0.32 \\
Unlabeled & GPT-5.4 & 0.99 & 0.97 & 0.94 & 0.37 & 0.13 \\
Unlabeled & Grok-4.1 & 0.95 & 0.97 & 0.96 & 0.63 & 0.29 \\
Unlabeled & Kimi-K3 & 0.98 & 0.99 & 0.98 & 0.31 & 0.12 \\
Unlabeled & Llama-4-Maverick & 0.94 & 0.97 & 0.98 & 0.15 & 0.11 \\
Unlabeled & Mistral-Large-3 & 1.00 & 0.98 & 0.99 & 0.31 & 0.12 \\
\midrule
Labeled & Human & 0.97 & 0.82 & 0.94 & 0.79 & 0.33 \\
Labeled & Claude-Opus-4.6 & 0.96 & 0.93 & 0.98 & 0.64 & 0.25 \\
Labeled & DeepSeek-V4-Pro & 0.95 & 0.96 & 0.98 & 0.28 & 0.15 \\
Labeled & Gemini-3.1-Pro & 0.98 & 1.00 & 0.99 & 0.94 & 0.31 \\
Labeled & GPT-5.4 & 0.99 & 0.96 & 0.94 & 0.42 & 0.18 \\
Labeled & Grok-4.1 & 0.97 & 0.98 & 0.96 & 0.58 & 0.29 \\
Labeled & Kimi-K3 & 0.98 & 1.00 & 0.99 & 0.35 & 0.17 \\
Labeled & Llama-4-Maverick & 0.96 & 0.95 & 0.97 & 0.14 & 0.06 \\
Labeled & Mistral-Large-3 & 0.98 & 0.96 & 0.99 & 0.29 & 0.13 \\
\bottomrule
\end{tabular}
\caption{Same-winner rate between each rubric dimension and overall preference. Ties on either dimension are excluded, and each entry reports how often the two questions select the same version.}
\label{tab:dimension-overall-association-app}
\end{table}

\subsection{Human Trade-off and Click-Likelihood Model}
\label{app:human-click-model}

We further quantify the human trade-off behind the Pareto view in \Cref{fig:human-pareto-frontier} in \Cref{tab:price-of-prominence-app}. Moving an advertisement earlier often increases visibility, but the gain is accompanied by a loss in overall preference. The end-to-front move is especially revealing. In the unlabeled condition, visibility increases by $0.666$, while overall preference decreases by $0.150$. In the labeled condition, the visibility gain is smaller, $0.514$, and the overall-preference loss is larger, $0.188$. Labeling therefore changes both the preferred placement and the cost of using position to obtain salience.
\begin{table}[!ht]
\centering
\setlength{\tabcolsep}{5pt}
\renewcommand{\arraystretch}{1.04}
\begin{tabular}{llrrr}
\toprule
Condition & Move & $\Delta$(Visibility) & $\Delta$(Overall) & Cost per visibility gain \\
\midrule
Labeled & End $\rightarrow$ Late & $+0.014$ & $-0.112$ & $-8.00$ \\
Labeled & Late $\rightarrow$ Early & $+0.095$ & $-0.043$ & $-0.45$ \\
Labeled & Early $\rightarrow$ Front & $+0.405$ & $-0.033$ & $-0.08$ \\
Labeled & End $\rightarrow$ Front & $+0.514$ & $-0.188$ & $-0.37$ \\
\midrule
Unlabeled & End $\rightarrow$ Late & $-0.119$ & $+0.070$ & $-0.59$ \\
Unlabeled & Late $\rightarrow$ Early & $+0.029$ & $-0.118$ & $-4.07$ \\
Unlabeled & Early $\rightarrow$ Front & $+0.756$ & $-0.102$ & $-0.14$ \\
Unlabeled & End $\rightarrow$ Front & $+0.666$ & $-0.150$ & $-0.23$ \\
\bottomrule
\end{tabular}
\caption{Human placement-score changes when the advertisement is moved earlier. Cost per visibility gain is $\Delta$(Overall) divided by $\Delta$(Visibility), so negative values indicate overall-preference loss per unit of visibility gain.}
\label{tab:price-of-prominence-app}
\end{table}

We finally examine how stated click likelihood relates to the two dimensions most directly involved in the placement trade-off. For each human comparison, we encode whether Version A wins visibility and whether Version A wins overall preference as $x_{\mathrm{vis}},x_{\mathrm{ovr}}\in\{-1,0,+1\}$, where $+1$ means Version A wins, $-1$ means Version B wins, and $0$ means Tie. We then fit a Davidson-Bradley-Terry model for the click-likelihood choice,
\begin{align}
\Pr(\text{click Version A}) &=
\frac{\exp(\eta)}{\exp(\eta)+1+\nu\exp(\eta/2)}, \nonumber \\
\Pr(\text{click Version B}) &=
\frac{1}{\exp(\eta)+1+\nu\exp(\eta/2)}, \nonumber \\
\Pr(\text{Tie}) &=
\frac{\nu\exp(\eta/2)}{\exp(\eta)+1+\nu\exp(\eta/2)},
\end{align}
with
\begin{equation}
\eta = \beta_{\mathrm{vis}}x_{\mathrm{vis}}+\beta_{\mathrm{ovr}}x_{\mathrm{ovr}}+\delta .
\end{equation}
Here $\beta_{\mathrm{vis}}$ and $\beta_{\mathrm{ovr}}$ measure how strongly visibility and overall preference are associated with the stated click-likelihood choice. The term $\delta$ captures residual preference for Version A, and $\nu$ is the Davidson tie parameter. We fit the model separately for the unlabeled and labeled conditions.

\Cref{tab:click-vis-overall-app} shows that both visibility and overall preference are associated with stated click likelihood. The two coefficients are positive in both disclosure conditions, and likelihood-ratio tests against nested models indicate that each dimension contributes information beyond the other. The overall-preference coefficient is larger than the visibility coefficient in both conditions, suggesting that click likelihood in this benchmark is not simply a salience judgment. This supports the point-estimate trade-off in \Cref{fig:human-pareto-frontier} that placements must be evaluated jointly on visibility, click likelihood, and user preference.

\begin{table}[!ht]
\centering
\setlength{\tabcolsep}{4pt}
\renewcommand{\arraystretch}{1.05}
\begin{tabular}{llc}
\toprule
Condition & Term & Coef.\ [95\% CI] \\
\midrule
Labeled & Visibility & $+1.30$ {\scriptsize[+1.13, +1.53]} \\
Labeled & Overall preference & $+2.15$ {\scriptsize[+1.98, +2.33]} \\
\cmidrule(lr){2-3}
Labeled & Version-A term $\delta$ & $-0.011$ {\scriptsize[-0.179, +0.167]} \\
Labeled & Tie parameter $\nu$ & $+0.568$ {\scriptsize[+0.506, +0.639]} \\
\midrule
Unlabeled & Visibility & $+1.24$ {\scriptsize[+1.06, +1.42]} \\
Unlabeled & Overall preference & $+1.93$ {\scriptsize[+1.77, +2.12]} \\
\cmidrule(lr){2-3}
Unlabeled & Version-A term $\delta$ & $+0.038$ {\scriptsize[-0.082, +0.171]} \\
Unlabeled & Tie parameter $\nu$ & $+0.428$ {\scriptsize[+0.367, +0.483]} \\
\bottomrule
\end{tabular}
\caption{Davidson-Bradley-Terry model for human stated click likelihood. Confidence intervals are cluster-bootstrap intervals over source instances with $B=200$ resamples. Likelihood-ratio tests compare the full model with nested models omitting one predictor. Adding visibility to a model with overall preference gives $\chi^2(1)=187$ for unlabeled and $\chi^2(1)=174$ for labeled; adding overall preference to a model with visibility gives $\chi^2(1)=554$ and $\chi^2(1)=677$. All four tests have $p<0.001$.}
\label{tab:click-vis-overall-app}
\end{table}

\end{document}